%% file: main.tex
\documentclass[lettersize,journal]{IEEEtran}
\usepackage{hyperref}
\hypersetup{
    colorlinks=true,
    linkcolor=blue,
    filecolor=magenta,      
    urlcolor=cyan,
    pdftitle={Overleaf Document},
}
\usepackage{graphicx}
\usepackage{subcaption}
\usepackage{amsmath,amsfonts,amssymb}
\usepackage{algorithm}
\usepackage{algpseudocode}
\usepackage{array}
\usepackage{textcomp}
\usepackage{stfloats}
\usepackage{url}
\usepackage{verbatim}
\usepackage{cite}
\usepackage{multirow}
\usepackage{booktabs} % For professional looking tables
\usepackage{tabularx} % For adjustable-width columns and automatic line wrapping
\usepackage[table]{xcolor} % For row highlighting in tables
\begin{document}

\title{Semantic Haptic Feedback Enhances Dexterous Robotic Teleoperation}

% \author{Bingjian Huang,~\IEEEmembership{~IEEE Student Member}
\author{Bingjian Huang, Sahar Aseeri, Jonas Schmidtler, Joseph Zhang, Sonny Chan, Andrew Doxon, Jom Preechayasomboon, Evan Pezent, Alberto Rigo, Amir Memar, Nicholas Colonnese, Chase Tymms

        % <-this % stops a space
\thanks{This paper was produced by the IEEE Publication Technology Group. They are in Piscataway, NJ.}% <-this % stops a space
\thanks{Manuscript received April 19, 2021; revised August 16, 2021.}}

\markboth{Journal of \LaTeX\ Class Files,~Vol.~14, No.~8, August~2021}%
{Shell \MakeLowercase{\textit{et al.}}: A Sample Article Using IEEEtran.cls for IEEE Journals}

\IEEEpubid{0000--0000/00\$00.00~\copyright~2021 IEEE}

\maketitle

\begin{abstract}
In robot teleoperation, haptic feedback can be used to help human operators accomplish dexterous manipulation tasks. However, existing haptic feedback methods try to replicate high-fidelity sensory haptics that are felt in real world interactions, which are constrained by the sensing and feedback hardware capability and may lead to higher workload.

To addresses these limitations, this work introduces semantic haptics for teleoperation, which uses abstract haptic patterns to convey critical information about robot states. We categorize robot states into "Confirmations" and "Exceptions", implement a modular haptic rendering pipeline in robot simulation, and deliver semantic haptic feedback to operators through pneumatic and vibrotactile wristbands. This simplifies hardware requirements and enables one-to-many mappings between haptic patterns and robot states. 

Through three evaluation studies, we identify the most effective semantic haptic design for a common pick and place teleoperation task and compare semantic haptics to other teleoperation feedback approaches including sensory haptics and visual feedback. Results suggest that while semantic haptics performs similarly as other feedback in unimanual tasks, it achieves superior performance in bimanual tasks, with reduced task workload, increased situational awareness, and overall preference.
\end{abstract}

\begin{IEEEkeywords}
Telerobotics and Teleoperation, Haptics and Haptic Interfaces
\end{IEEEkeywords}

\begin{figure*}
    \centering
    \includegraphics[width=0.9\linewidth]{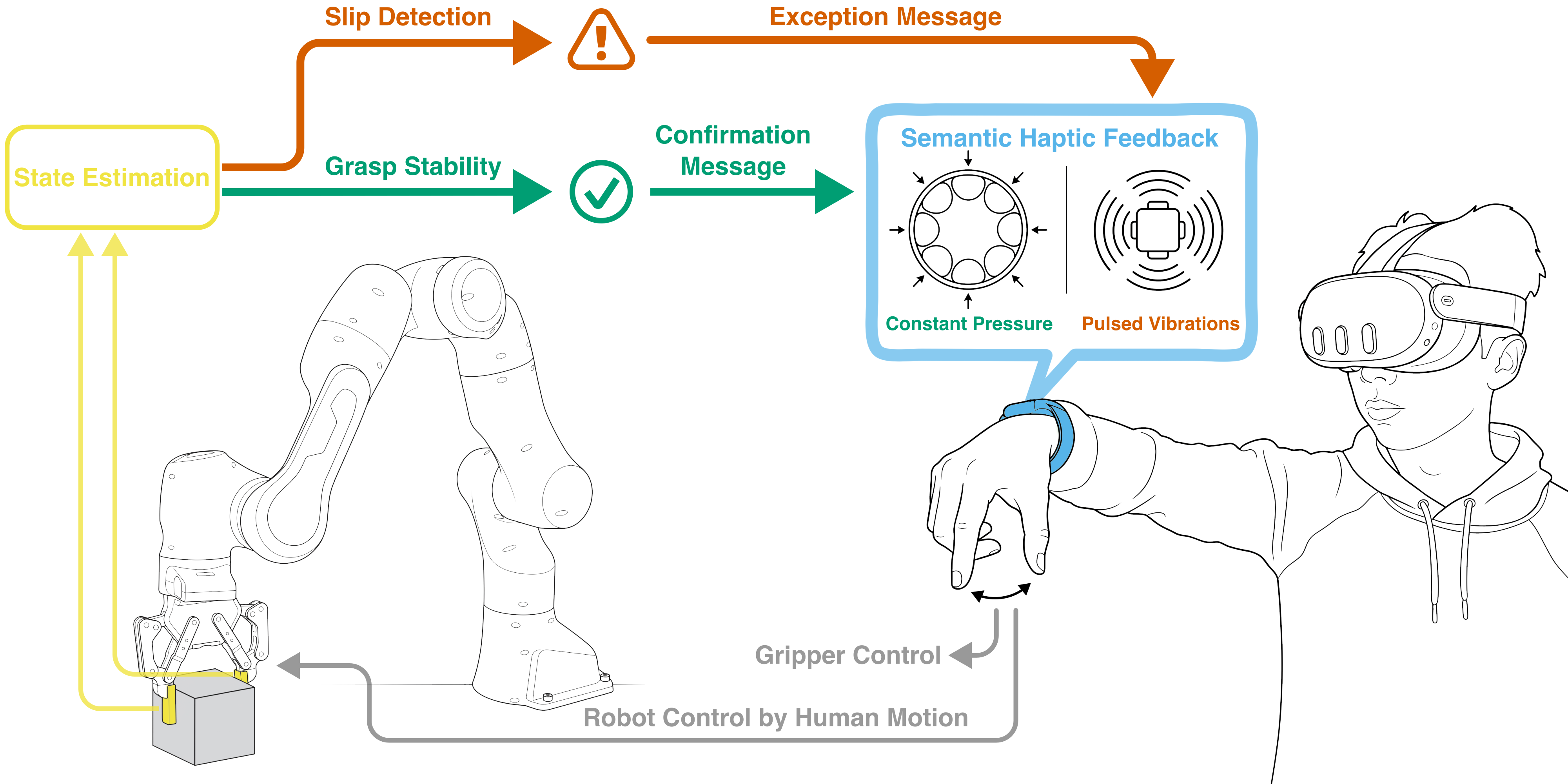}
    \caption{Semantic haptic feedback utilizes symbolic haptic patterns to communicate discrete, abstract information regarding critical robot and object states.}
    \label{fig:teaser}
\end{figure*}

\input{docs/1_introduction}

\input{docs/2_relatedwork}
\input{docs/3_systemoverview}
\input{docs/4_hapticstudy}
\input{docs/5_unimanualstudy}
\input{docs/6_bimanualstudy}
\input{docs/7_discussion}

% \section*{Acknowledgments}
% This should be a simple paragraph before the References to thank those individuals and institutions who have supported your work on this article. Joseph Zhang

% {\appendix[Proof of the Zonklar Equations]
% Use $\backslash${\tt{appendix}} if you have a single appendix:
% Do not use $\backslash${\tt{section}} anymore after $\backslash${\tt{appendix}}, only $\backslash${\tt{section*}}.
% If you have multiple appendixes use $\backslash${\tt{appendices}} then use $\backslash${\tt{section}} to start each appendix.
% You must declare a $\backslash${\tt{section}} before using any $\backslash${\tt{subsection}} or using $\backslash${\tt{label}} ($\backslash${\tt{appendices}} by itself
%  starts a section numbered zero.)}

\bibliographystyle{IEEEtran}
\bibliography{main}

\vfill

\end{document}

%% file: docs/1_introduction.tex
\section{INTRODUCTION}

% Robot teleoperation represents one essential way for collecting high-fidelity demonstration data for imitation learning and for operating in hazardous environments. However, teleoperation remains challenging because physical decoupling can severely degrade the operator's situational awareness. This limitation is critical in dexterous manipulation tasks, where humans rely on subtle tactile and kinesthetic cues to estimate object states and modulate contact forces. Without haptic feedback, precision tasks like handling fragile objects or performing precise insertions remain difficult to execute.

Robot teleoperation has emerged as a cornerstone of modern robotics, serving as both the primary method for collecting high-fidelity demonstration data for imitation learning~\cite{zhao2023learning,black2024pi0} and a critical tool for operating in hazardous or remote environments~\cite{murphy2014disaster,nagatani2013emergency,fishel2020tactile}. Despite its utility, teleoperation remains inherently challenging due to the physical decoupling of the operator from the robot, which severely degrades situational awareness~\cite{chen2007human} and task performance~\cite{triantafyllidis2020study, fishel2020tactile}. This limitation is particularly acute in dexterous manipulation tasks, where human performance relies on subtle tactile and kinesthetic cues to estimate object states and modulate contact forces~\cite{johansson2009coding}. Without sufficient haptic feedback, tasks requiring high precision and compliance, such as handling fragile objects and precise insertion, remain notoriously difficult to execute.

To address this, prior research explores various haptic modalities. Traditional approaches use haptic joysticks and exoskeletons to provide force feedback based on joint torques or end-effector workloads~\cite{horie2001remote, tang2009haptic, leonardis2024hand, lenz2021bimanual}. Recent developments focus on wearable tactile interfaces~\cite{pacchierotti2023cutaneous}. These devices employ vibrotactile actuators, pneumatic chambers, or pin arrays to convey granular contact data like geometry and alignment~\cite{kontarinis1995tactile, abd2018armband, colella2019novel, meli2014sensory}. Yet results remain mixed~\cite{pacchierotti2023cutaneous, khurshid2016effects, vandermeijden2009value}: force feedback is prone to instability under bandwidth and latency constraints~\cite{pacchierotti2015cutaneous, tanioka2025effects}, and while it consistently reduces applied force, it does not improve success rate, even for expert users~\cite{wagner2007benefit}.

% To bridge this gap, prior research has explored using various modalities of haptic feedback to assist teleoperation tasks. Traditional approaches utilize haptic joysticks and exoskeletons to provide force feedback mapped from joint torques or end-effector workloads [XXX]. More recent developments focus on wearable tactile interfaces~\cite{pacchierotti2023cutaneous} employing vibrotactile actuators, pneumatic chambers, and pin arrays to convey granular contact information, including contact onset, object geometry, in-hand orientation, and high-frequency acceleration [XXX]. However, these efforts have yielded mixed results [XXX]. Evidence suggests that the addition of haptic feedback does not monotonically improve task performance; in many cases, it increases cognitive load without a commensurate gain in efficiency, even among expert operators [XXX]

The primary bottleneck in current haptic teleoperation is a heavy reliance on \textbf{sensory haptic feedback}~\cite{pacchierotti2023cutaneous, culbertson2018haptics}. This approach aims to replicate realistic touch sensations by directly mapping low-level sensor data to haptic signals in a one-to-one manner (e.g., robot force sensors to operator pressure feedback). However, high-fidelity replication remains challenging because current hardware cannot match the spatial resolution and temporal bandwidth of human perception~\cite{tan2020methodology, choi2013vibrotactile, culbertson2018haptics}. The resulting fidelity gap can introduce perceptual artifacts and control instabilities that hinder rather than aid the operator~\cite{pacchierotti2015cutaneous, meli2014sensory}. Furthermore, because most interfaces only support single modality, achieving realistic multimodal feedback requires complex, bulky hardware~\cite{shtarbanov2023sleeveio, lin2025hcr}. Consequently, sensory haptic feedback is insufficient for scalable teleoperation platforms.

% We argue that a primary bottleneck in current haptic teleoperation lies in the heavy reliance on \textbf{sensory haptic feedback}, which aims to capture the realistic touch sensations from the robot and replicate them for the operator. To achieve such high haptic transparency, these works often directly map low-level sensor data to haptic signals in a one-to-one manner (e.g., normal force sensor on the robot to pressure feedback on the operator). However, the high-fidelity replication of physical interactions remains a challenging task, because current technical solutions cannot meet the spatial resolution and the temporal bandwidth requirements of human perception. As evidenced by similar haptics research in virtual reality community, it often creates a "sensation gap" that overwhelms the operator rather than aiding them [XXX]. Furthermore, since most haptic interfaces are designed for a single modality (e.g., vibrotactile actuators for vibrations, achieving realistic multimodal feedback requires complex, non-scalable hardware integration [SleeveIO, four modality handheld bulky device]. Consequently, direct physical replication is likely an insufficient methodology for developing robust, scalable teleoperation platforms.

To address these limitations, we propose a shift toward \textbf{semantic haptic feedback}. Instead of replicating physical touch, this approach uses symbolic haptic patterns to communicate discrete, abstract state information~\cite{maclean2003perceptual,brewster2004tactons}. For example, during a pick-and-place task, rather than streaming continuous gripping force data which can be mentally demanding, our system estimates grasp stability and triggers brief patterns to confirm a "stable grasp" or warn the exception of an "incipient slip". Operators do not need exhaustive raw data. Instead, they benefit more from high-level updates on state transitions to guide motion planning~\cite{johansson2009coding}.

% To address these limitations, we propose a paradigm shift toward \textbf{semantic haptic feedback}. Rather than focusing on replication of realistic touch, this approach utilizes symbolic haptic patterns [Maclean, XXX] to communicate discrete, abstract information regarding critical robot and object states. For instance, in a pick-and-place task, instead of streaming continuous gripping force data (which is mentally demanding to process), our approach evaluates grasp pose and triggers specific haptic patterns to confirm a "stable grasp" or warn a "incipient slip". This methodology is predicated on the insight that operators do not require exhaustive contact data; instead, they benefit more from high-level updates on object state transitions, allowing for more effective high-level motion planning.

Semantic haptic feedback offers three key advantages over traditional sensory feedback methods:
\begin{itemize}
\item \textit{Generalized One-to-Many Mapping}: A single haptic primitive can represent various robot states across different tasks. For example, a wrist squeeze can signify "grasp confirmation" during picking or "alignment success" during insertion. This sharing reduces the user's learning curve.
\item \textit{Simplified Hardware Requirements}: Decoupling feedback from strictly matching haptic modality allows effective communication through low-degrees-of-freedom (low-DoF) hardware, like standard vibrotactile wristbands. This facilitates scalable deployment and data collection.
\item \textit{Reduced Workload}: Transitioning to an event-driven model frees the operator from continuously monitoring raw signals. This improves multimodal attention allocation and enhances situational awareness during dexterous tasks.
\end{itemize}

% Semantic haptic feedback offers three primary advantages over traditional analog methods:
% \begin{itemize}
%     \item Generalized One-to-Many Mapping: By employing symbolic patterns, a single haptic primitive can represent robot states across diverse tasks. For example, a squeeze on the wrist can signify "grasp confirmation" in pick-and-place or "orientation alignment" in peg-in-hole insertion. This reduces the operator's learning curve, as a small set of cues can be reused for various manipulation tasks.
%     \item Simplified Hardware Requirements: Decoupling the feedback from physical modality allows for effective communication even through low-degrees-of-freedom (low-DoF) hardware, such as vibrotactile or pneumatic wristbands. This significantly lowers the barrier for large-scale data collection and scalable deployment.
%     \item Reduced Cognitive Load and Improved Situation Awareness: By transitioning to an event-driven feedback model, the operator is freed from the "proprioceptive burden" of monitoring continuous signals. This leads to a more balanced distribution of multimodal attention and improves operator's situation awareness in dexterous manipulation tasks.
% \end{itemize}

We developed an end-to-end simulated teleoperation pipeline to explore semantic haptic feedback. The pipeline simulates a bimanual dexterous manipulation platform in an interactive VR game engine and derives real-time 3-axis force at contact points. This contact-rich information is fed into state estimation algorithms like grasp stability and slip detection. When critical robot state changes are detected, these events are delivered to operators as multimodal haptic feedback, via integrated wearable devices with vibrotactile and pneumatic actuators. Using this pipeline, we aim to address two research questions:
\begin{enumerate}
    \item \textit{How can semantic haptic feedback be designed to optimize teleoperation performance across different manipulation tasks?}
    \item \textit{How does operator performance under semantic haptic feedback compare to that under visual or sensory haptic baselines?}
\end{enumerate}

To address the first question, we evaluated four combinations of vibrotactile and pneumatic feedback in an assisted unimanual pick-and-place task. Performance was optimized when pneumatic cues provided confirmation and vibrotactile cues provided warnings. This indicates that semantic designs should align with the inherent meaning of each modality, and multimodal feedback is more effective than single modality.

To answer the second question, we evaluated performance across both the original unimanual task and an expanded bimanual setting. We compared four conditions: a no-feedback baseline, semantic visual feedback, sensory haptic feedback, and semantic haptic feedback. While performance was similar in the unimanual task due to ample visual feedback, semantic haptic feedback significantly outperformed the other conditions in the bimanual task and was strongly preferred by participants, with reduced workload and increased situational awareness. Qualitative feedback confirmed that operators felt less overwhelmed and more confident during bimanual operations. We conclude by discussing design implications and future directions.

%% file: docs/2_relatedwork.tex
\section{Related Work}

\subsection{Haptic Feedback in Teleoperation}

Teleoperation enhances performance in safety-critical domains like surgery, construction, and disaster response~\cite{niemeyer2016telerobotics}, and is increasingly utilized for robot policy data collection. However, relying primarily on visual feedback limits the operator's situational awareness during physical interactions. Without direct physical cues, executing dexterous manipulation tasks such as regulating grasping forces for fragile objects or aligning components during assembly remains highly challenging. This ultimately degrades teleoperation performance and limits the range of tasks possible.

% Teleoperation has been shown to increase safety and comfort for human
% workers in applications such as minimally invasive surgery, search and rescue,
% and construction~\cite{niemeyer2016telerobotics}. Recently, it has also been increasingly used in data collection for robot policy training. In such scenarios,
% operators rely predominantly on direct vision or camera feeds to
% operate the robot. Because of the physical separation between the robot and the human operator, the operator receives limited cues from robot interaction and thus finds it difficult to perform dexterous manipulation tasks. For example, when picking up a fragile object, it is hard to tell how much force is needed to pick up the object without breaking it; when inserting a USB cable, it is hard to tell whether the USB header is aligned with the female socket or not before insertion. This lack of situational awareness has dampened the operator performance and limited the range of dexterous manipulation tasks possible in teleoperation.

To address this, researchers integrate haptic feedback into teleoperation pipelines, broadly categorizing methods into kinesthetic force and cutaneous tactile feedback~\cite{pacchierotti2023cutaneous}. Force feedback methods measure contact forces using internal force/torque sensors and render them via grounded joysticks, exoskeletons, or leader-follower structures. For instance, force cues have been mapped to grounded devices for end-effector control~\cite{horie2001remote} and heavy machinery operation~\cite{tang2009haptic}. Arm and hand exoskeletons have also been developed to render joint torques and fingertip forces~\cite{leonardis2024hand, lenz2021bimanual}, while custom handheld controllers approximate specific hand shapes or gripper configurations to handle fragile objects~\cite{zhang2025doglove, satsevich2025prometheus}. Despite improving grasping safety and user performance~\cite{tang2009haptic, leonardis2024hand}, force feedback systems face limited industrial adoption due to high hardware costs and potential control instabilities induced by communication latency; even minimal delays (e.g., 20--100~ms) can severely degrade task efficiency and operability~\cite{pacchierotti2023cutaneous, tanioka2025effects, meli2014sensory}.

Alternatively, cutaneous and tactile feedback mitigates stability issues by decoupling control inputs from haptic outputs in an open-loop configuration~\cite{pacchierotti2023cutaneous}. By leveraging "sensory subtraction," researchers isolate destabilizing force feedback while preserving necessary interaction cues via fingertip skin deformation devices~\cite{meli2014sensory, pacchierotti2015cutaneous}. Subsequent designs have utilized multi-bar linkages to render complex skin stretch, slip, and twist patterns~\cite{zhu2022cutaneous}, or armbands to deliver normal and shear forces~\cite{colella2019novel, casini2015design}. For vibrotactile feedback, early work mapped force magnitudes to fixed-frequency vibrations~\cite{massimino1993sensory}, while modern approaches utilize off-the-shelf VR controllers paired with force or vision-based tactile sensors~\cite{kamijo2024learning, lippi2024low}, or deliver naturalistic tool vibrations directly to operator handles~\cite{kuchenbecker2010verrotouch}. To replicate rich interaction sensations, recent trends explore multimodal devices that combine grip force, contact pressure, and vibrations through ungrounded handles~\cite{khurshid2016effects}, pneumatic-vibration mechanisms~\cite{abd2018armband}, or multi-cue continuum robots~\cite{lin2025hcr}.

Although advanced tactile feedback designs provide rich sensations, scaling them to complex, multi-state robotic tasks remains difficult. Adding distinct feedback modalities exponentially increases hardware cost, system integration complexity~\cite{lin2025hcr}, and user cognitive workload. Consequently, design principles for a simple, effective, and adaptable haptic interface suitable for scalable teleoperation pipelines remain an open question.

\begin{table*}[!t]
\caption{Tactile feedback modalities, input sensors, output devices, and mappings in robot arm teleoperation, grouped by modality. Papers spanning multiple modalities appear in each applicable row. The last row summarizes the semantic haptic approach proposed in this work.}
\label{tab:tactile_teleop_review}
\centering
\renewcommand{\arraystretch}{1.25}
\begin{tabularx}{\textwidth}{@{}>{\raggedright\arraybackslash}p{3.1cm} X X X@{}}
\toprule
\textbf{Modality} & \textbf{Input Sensors} & \textbf{Output Devices} & \textbf{Common Mappings} \\
\midrule
Indentation~\cite{kontarinis1995tactile, chang1999kist, shen2003haptic, sarakoglou2012high, fishel2020tactile, king2009tactile, uddin2016pneumatic, park2016force, abd2018armband, pacchierotti2015cutaneous, khurshid2016effects, clark2019role}
& Force/pressure sensors; tactile arrays (BioTac, FlexiForce); contact-area estimators
& Pin arrays; pneumatic chambers; motor- or cable-driven fingertip platforms; MR-fluid pads
& Contact force or pressure $\rightarrow$ indentation depth or normal pressure \\
\addlinespace[2pt]
Lateral motion~\cite{roke2013effects, quek2019evaluation, zhu2022cutaneous, clark2019role, fani2018simplifying}
& Shear/tangential force sensors; end-effector F/T sensors; object pose
& Laterally-shifted pin arrays; skin-deformation platforms; arm cuffs (e.g., CUFF)
& Shear force or object rotation $\rightarrow$ skin stretch, slip, or twist \\
\addlinespace[2pt]
Vibration~\cite{khurshid2016effects, watanabe1995method, tokashiki2000effects, tsetserukou2009teleta, martinez2017towards, hayashi2009teleoperation, mcmahan2010high}
& Accelerometers; contact-event detectors; texture estimators
& Voice-coil actuators on handles, wristbands, or gloves; ultrasonic plates
& Contact events $\rightarrow$ vibration bursts; force or texture amplitude $\rightarrow$ vibration intensity \\
\addlinespace[2pt]
Kinesthetic~\cite{chang1999kist, sarakoglou2012high, fishel2020tactile, khurshid2016effects, quek2019evaluation, zhu2022cutaneous, tokashiki2000effects, tsetserukou2009teleta}
& End-effector F/T sensors; joint torque sensors; robot pose
& Grounded joysticks (Omega 7, da Vinci MTM); exoskeletons; leader--follower master arms
& End-effector force or joint torque $\rightarrow$ resistive force on the hand or arm \\
\addlinespace[2pt]
Electrotactile~\cite{sato2007improvement}
& Contact-area sensing on robot finger
& Electrotactile electrode arrays on the fingertip
& Contact-area pattern $\rightarrow$ electrode activation pattern \\
\midrule
\rowcolor{gray!15}
\textbf{Semantic haptics (ours)}: pneumatic + vibrotactile
& 3-axis end-effector contact forces from a physics estimated from a physics simulator: grasp stability and slip detection.
& Bilateral wrist-mounted wearables with pneumatic cells and vibrotactile actuators, located away from the contact site
& Cross-modal, one-to-many: robot states $\rightarrow$ custom haptic patterns encoding phase confirmations and subgoal exceptions \\
\bottomrule
\end{tabularx}
\end{table*}

\subsection{Semantic Haptic Design}

Rather than replicating physical sensations, semantic haptic design uses tactile cues to convey abstract task and system states. Maclean~\cite{maclean2003perceptual} formalized this method as using haptic feedback to communicate abstract messages: \textit{"The bulk of applications for haptic feedback employ direct rendering approaches wherein a user touches a virtual model of some “real” thing… We propose a new class of applications based on abstract messages, ranging from “haptic icons” – brief signals conveying an object’s or event’s state, function or content – to an expressive haptic language for interpersonal communication."}

% Instead of trying to replicate the realistic touch sensations, the authors proposed using semantic haptic design to effectively convey robot state information. Semantic haptic design describes the method of using haptic / tactile feedback to communicate abstract messages of task and system statuses to the user. As suggested by Maclean~\cite{maclean2003perceptual}, \textit{"The bulk of applications for haptic feedback employ direct rendering approaches wherein a user touches a virtual model of some “real” thing… We propose a new class of applications based on abstract messages, ranging from “haptic icons” – brief signals conveying an object’s or event’s state, function or content – to an expressive haptic language for interpersonal communication."} 

As an alternative to visual and auditory communications, semantic haptics effectively transmits complex spatiotemporal patterns. In gesture interaction, Xu et al.~\cite{xu2024designing} mapped metaphoric vibrotactile patterns to gesture input commands, significantly improving interaction efficiency. For the deaf community, Reed et al.~\cite{reed2018phonemic} developed a 24-actuator sleeve that transmitted 39 phonemes with an 86\% recognition rate. In affective computing, Rognon et al.~\cite{rognon2022linking} successfully used a haptic glove to communicate social touch messages via five patterns. %%%more papers needed

% Semantic haptics serves as an communication channel alternative to visual and audio channels, and have demonstrated its strength in conveying spatiotemporal patterns in applications such as gesture interaction, accessibility and affective computing. For instance, Xu et al~\cite{xu2024designing} designed nice different vibrotactile patterns and mapped them to gesture input commands using metaphors. Study results showed that vibrotactile patterns help users confirm their input commands right after each gesture, leading to more efficient interactions. Reed et al~\cite{reed2018phonemic} built a tactile sleeve with 24 actuators and designed spatiotemporal patterns to deliver 39 phonemes. Study results showed a mean recognition rate of 86 percent correct within one to four hours of training across participants, thus proving the viability of conveying speech information through semantic haptics. Rognon et al.~\cite{rognon2022linking} investigated five haptic patterns to communicate tactile messages through social touch communicated via a haptic glove. %%%more papers needed

Designing these patterns requires maximizing communication bandwidth while minimizing user cognitive workload~\cite{tan2020methodology,pasquero2006perceptual}. Metaphor-based approaches leverage a user's prior auditory, visual, or associative experiences to lower interpretation effort~\cite{xu2024designing,sung2025hapticgen}. This process is often supported by established design guidelines and tacton databases~\cite{brewster2004tactons,seifi2015vibviz}. Alternatively, affective touch designs vary haptic parameters to evoke specific valence and arousal~\cite{yohanan2012role,rognon2022linking,cang2023haptic,vyas2023descriptive}. This allows users to intuitively recognize messages based on familiar social gestures.

% Designing semantic haptic patterns remain a challenging job, as patterns need to afford high communication bandwidth while balancing the cognitive load of memorization and recognition. In prior work, popular design strategies include metaphors and affective touch. Metaphors associated with haptic patterns are mental models that harness users’ pre-existing experiences—such as audio [XXX HapticGen, Xu], visual experiences, or other associations—with their encoded meanings to lower the cognitive load necessary to understand, remember, and retain their patterns. Prior work has also proposed pre-designed tacton databases [XXX] and design guidelines to aid in the process of designing metaphor-based haptic patterns. On the other hand, affective touch design utilizes common social touch gestures with inherent meanings. As emotional feelings of valence and arousal have been reported to be evoked by haptic signals with varying haptic parameters, encoded emotional messages could be intuitively recognized if haptic patterns trigger emotional feelings that match the messages.

Despite the prevalence of semantic haptics in interaction design, it has only recently been applied to robotics, primarily for teleoperation and learning from demonstration. Huang et al.~\cite{huang2025aerohaptix} modulated vibration amplitudes on a haptic jacket to communicate obstacle proximity during drone teleoperation. Valdivia et al.~\cite{valdivia2023wrapping, valdivia2026modular} integrated a pneumatic display onto a robotic arm to communicate task confidence and request human intervention. However, to the best of our knowledge, semantic haptics remains unexplored in robotic dexterous manipulation. This work introduces the first semantic haptic pipeline for robot arm teleoperation and evaluates its impact on operator performance during dexterous tasks.

\subsection{Definition of Semantic Haptics}
% clearly define semantic haptics to distinguish it from sensory haptics which tries to relicate physical sensations.

To clearly distinguish our work from prior haptic robot teleoperation work, we provided a \textbf{formal definition of semantic haptics} in the context of robot teleoperation:

\textit{Semantic haptic feedback refers to the usage of various modality of haptic feedback with custom designed patterns to communicate abstract messages of critical robot states to the robot teleoperator, for the purpose of improving task performance and reducing task workload.}

And the key features that distinguish semantic haptics from traditional sensory haptics are summarized in Table \ref{tab:sec2_haptics_twocolumn}. Specifically, semantic haptics is not directly mapped to sensory data or trying to replicate physical sensations. As a counter example, using air pressure to replicate the realistic sensations of pressing or grasping an object is a form of sensory haptics. Additionally, the feedback location can be different from the location where physical contacts take effect. For example, if a robot hand is touching an object, sensory haptic feedback needs to be rendered on the human hand, where as semantic haptic feedback can be rendered on other positions as well, such as wrist, arm and body trunk, etc.

\begin{table}[h!]
    \centering
    \caption{Sensory vs. Semantic Haptics}
    \label{tab:sec2_haptics_twocolumn}
    \vspace{0.5em}
    \footnotesize % Scales table size safely for a narrow column
    \begin{tabularx}{\columnwidth}{l >{\raggedright\arraybackslash}X >{\raggedright\arraybackslash}X}
        \toprule
        \textbf{Category} & \textbf{Sensory} & \textbf{Semantic} \\ 
        \midrule
        \textbf{Goal}     & Replicate physical sensations & Convey abstract messages \\ 
        \addlinespace
        \textbf{Source}   & Raw physical data             & Robot state estimation \\ 
        \addlinespace
        \textbf{Location} & At contact point               & Anywhere on body \\ 
        \addlinespace
        \textbf{Signal}   & Continuous tracking           & Discrete events and custom patterns \\ 
        \addlinespace
        \textbf{Mapping}  & Same modality                  & Cross-modal \\ 
        \bottomrule
    \end{tabularx}
\end{table}

% reference Choi's definition of tactile icons

%% file: docs/3_systemoverview.tex
\section{System Overview}
\label{sec3_systemoverview}
%TODO: debating if I should also include descriptions of the overall teleoperation framework so readers have a better idea of where the haptic pipeline sits in the framework.

\begin{figure*}
    \centering
    \includegraphics[width=\linewidth]{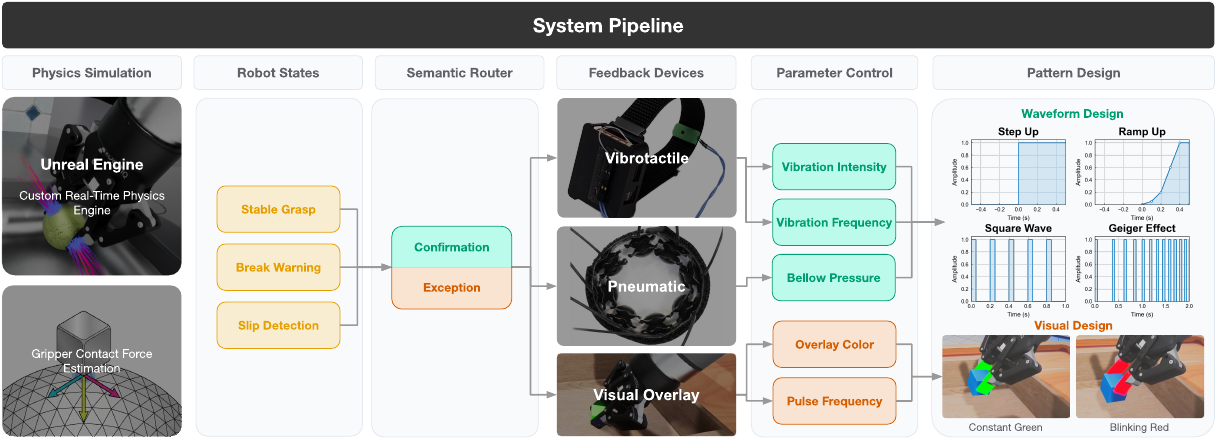}
    \caption{Semantic haptic teleoperation pipeline. Real-time physics engine feeds fingertip contact-rich information to state estimation algorithms to detect critical events of robot state changes. A semantic router maps the events into \textit{confirmation} and
\textit{exception} messages sent to multimodal pneumatic
and vibrotactile wristbands, with custom waveforms such as ramp-up and "Geiger Effect". A visual overlay on the gripper serves as a baseline.}
    \label{fig:sec3_systempipeline}
\end{figure*}

To explore the usage of semantic haptic feedback in dexterous manipulation, we built a haptic rendering pipeline in simulation environment. Our pipeline utilizes a reconfigurable design that transforms various robot states into unique semantic haptic patterns delivered via wearable tactile devices (Figure \ref{fig:sec3_systempipeline}). Because our goal is to build a haptic pipeline to evaluate the effectiveness of semantic haptic, we chose to build it in simulation to simplify state estimation and allows us to focus on testing haptic design. Below we will describe each part of the pipeline in details.

\subsection{Simulation Environment}

\begin{figure*}
    \centering
    \includegraphics[width=\linewidth]{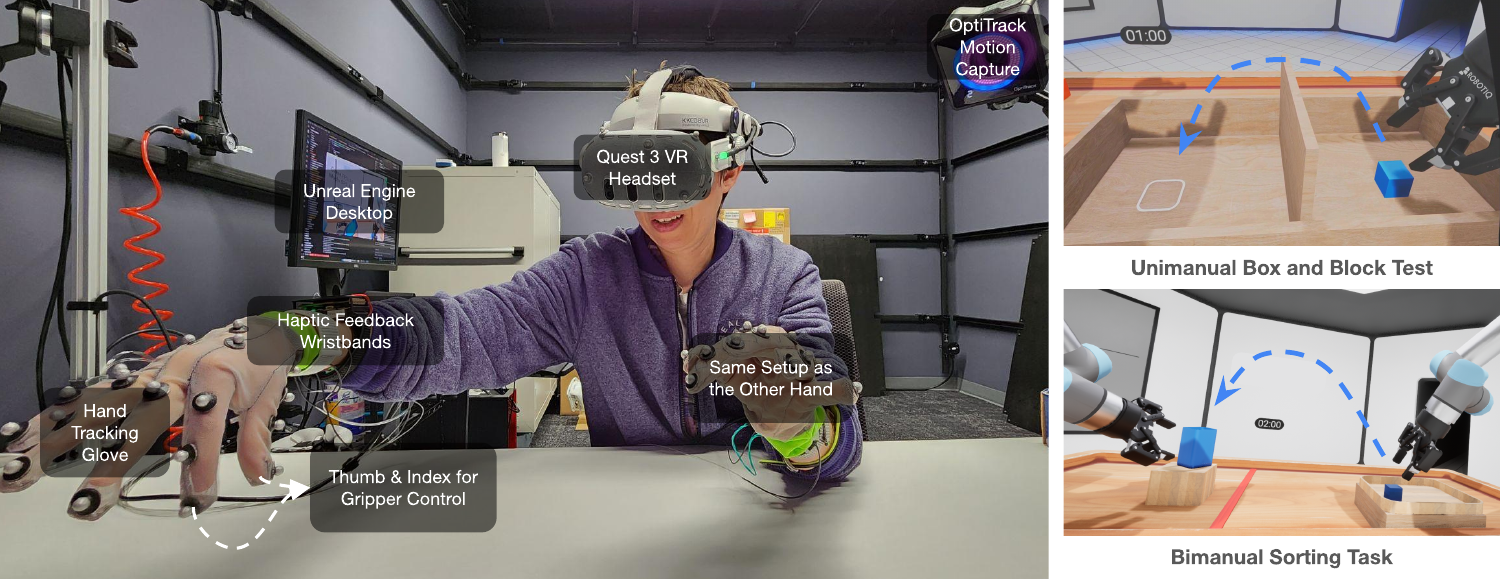}
    \caption{Teleoperation setup and interaction scenes built using the simulation environment for the evaluation studies.}
    \label{fig:sec3_scenes}
\end{figure*}

The system runs in a custom virtual robot teleoperation platform (Figure \ref{fig:sec3_scenes}) built with Unreal Engine 5. Robotic arms such as Franka Research 3 Arm\footnote{\url{https://franka.de/research}} and end-effectors such as Robotiq 2F-85 gripper\footnote{\url{https://robotiq.com/products/adaptive-grippers}} and various robot hands are imported into the platform using URDF files to represent the actuated joint structure. Various scene and objects are also created to support daily interaction tasks. During teleoperation, the operator wears motion tracking gloves tracked by OptiTrack\footnote{\url{https://www.optitrack.com/}} cameras to control the robot arm. The end effector of the arm follows the operator's wrist position and rotation, while the gripper clutch is controlled by the pinch distance between the thumb and index fingers. When the gripper grasps objects, the rich contact information between robots and objects are derived from the real-time computation of an in-house custom real-time physics engine. It computes the joint torques of robot arms and 3-axis force at the contact area between objects and grippers, providing ground-truth information for state estimation.

Utilizing the teleoperation platform, we built two interaction scenes (Figure \ref{fig:sec3_scenes}. The Box and Block Test~\cite{mathiowetz1985adult} measures unilateral force modulation dexterity, wherein the subject is asked to teleoperate a Franka FR3 Arm with Robotiq 2F-85 gripper (right handed) to pick up a cube from a box, move it across the partition, and place it into another box. While this task is relatively easy for healthy participants in real world, it is quite challenging in teleoperation as participants lose their intuitive sense of the grasping force. Fishel et al.~\cite{fishel2020tactile} have tested Box and Block and found that a human subject could move an average of 87.33 blocks in 1 minute in real world, but could only move 19 blocks in teleoperation, which was approximately 4.6 times slower. We make the task even more challenging by introducing fragile cubes. These cubes were designed with the fracture mode in Unreal Engine and were configurated to break if the contact force is above a certain threshold (e.g., 120N). Therefore, in practice participants need to modulate their grasp force very carefully, because too little force will cause the cube to slip and too much force will cause it to break.

The bimanual sorting task measures bimanual dexterity and attention switching, wherein the subject is asked to pick up a container using the left gripper, pick up a cube using the right gripper and place it into the container, then repeat the task over and over again. This task utilizes the same cube from the Box and Block test and adds a slippery container that is difficult to grasp. During interaction, when the user is paying close attention to the right gripper to pick up the cube, the left gripper might get loose or tight due to subconscious hand movement. Without proper haptic feedback, the container could slip out of the left gripper without the user noticing it. Thus, this task requires frequent attention switching and multimodal feedback could potentially be useful in this case.

\subsection{Robot State Estimation}

%%% TODO: remove this redundant figure

% \begin{figure}[htbp]
%     \centering
%     \begin{subfigure}{\columnwidth}
%         \centering
%         \includegraphics[width=\linewidth]{figures/sec3_gripperforcedetector.png}
%         \caption{Gripper Force Detector}
%         \label{fig:sec3_gripperforcedetector}
%     \end{subfigure}
    
%     \vspace{12pt}
    
%     \begin{subfigure}{\columnwidth}
%         \centering
%         \includegraphics[width=\linewidth]{figures/sec3_grippercontactsslipdetector.png}
%         \caption{Gripper Slip Detector}
%         \label{fig:sec3_grippercontactsslipdetector}
%     \end{subfigure}
    
%     \caption{Diagrams showing the working principles of the state estimation algorithms.}
%     \label{fig:sec3_algorithm_diagram}
% \end{figure}

We implemented multiple robot and object state estimation algorithms to assist human operators with the aforementioned dexterous manipulation tasks:
\begin{itemize}
    \item \textit{Gripper Force Detector}: Algorithm \ref{alg:force_detector} is attached to the right gripper in Unreal Engine and subscribes to the fingertip contact events reported by the physics layer. Each tick, the summed contact-force magnitudes from both fingertips are aggregated into $F$. Once $F$ exceeds the grasp threshold $T_{grasp}$, a confirmation feedback is triggered to notify the user that a stable grasp is established; once $F$ further exceeds the break threshold $T_{break}$, a warning feedback is emitted with intensity linearly ramped as $F$ approaches the maximum reference force $F_{max}$.
    \item \textit{Gripper Slip Detector}: Algorithm \ref{alg:slip_detector} is attached to the left gripper in Unreal Engine and subscribes to the fingertip contact events reported by the physics layer. Once a stable grasp is established, the detector monitors the total number of fingertip contact points $N$ between the gripper and the container. Under nominal contact, a stable grasp sustains a characteristically high $N$; when the applied force is either too strong or too weak, the container gradually slips from the gripper and $N$ drops sharply. We empirically identify two regime boundaries: a slip-onset threshold $N_{loss}$, below which slip is declared, and a release threshold $N_{min}$, at which the container has effectively left the gripper. The warning effect is triggered once $N$ falls below $N_{loss}$, with intensity linearly ramped as $N$ approaches $N_{min}$.

\end{itemize}

\begin{algorithm}[htpb]
\caption{Gripper Force-Based Grasp and Break Detection}
\label{alg:force_detector}
\begin{algorithmic}[1]
\Require Grasp threshold $T_{grasp}$; break threshold $T_{break}$ ($T_{break} > T_{grasp}$); maximum reference force $F_{max}$.
\Statex \textbf{Per-finger cache} (set by callback, cleared each tick): forces $f_R, f_L$; contacted actors $a_R, a_L$.
\Statex
\Procedure{OnNestedContact}{link, $\mathcal{E}_{contact}$}
    \State $\mathcal{C} \gets \textsc{FilterByTag}(\mathcal{E}_{contact}, \textit{``LeftFingertip''} \lor \textit{``RightFingertip''})$
    \State Update $(f_\ast, a_\ast)$ for the fingertip \textit{link} from $\mathcal{C}$
\EndProcedure
\Statex
\Procedure{Tick}{$\Delta t$}
    \State $F \gets f_R + f_L$ \Comment{Aggregated grasp force}
    \If{$F > T_{grasp}$} \Comment{Stable grasp confirmation}
        \State \text{TriggerGraspFeedback}()
    \EndIf
    \If{$F > T_{break}$} \Comment{Break warning}
        \State $\mathcal{I} \gets \mathrm{clip}\!\left(\dfrac{F - T_{break}}{F_{max} - T_{break}},\, 0,\, 1\right)$
        \State \text{TriggerBreakFeedback}($\mathcal{I}$)
    \EndIf
    \State Clear per-finger cache $(f_\ast, a_\ast)$
\EndProcedure
\end{algorithmic}
\end{algorithm}

\begin{algorithm}[htpb]
\caption{Gripper Contact-Based Slip Detection}
\label{alg:slip_detector}
\begin{algorithmic}[1]
\Require Grasp force threshold $T_{grasp}$; slip threshold $N_{loss}$; minimum-contact threshold $N_{min}$ ($N_{min} < N_{loss}$). Thresholds are set by the gripper geometry.
\Statex \textbf{Per-finger cache} (set by callback, cleared each tick): counts $c_R, c_L$; forces $f_R, f_L$; contacted actors $a_R, a_L$.
\Statex \textbf{Persistent state:} grasped actor $A_{grasp} \gets \varnothing$;\; $\textit{isStable} \gets \text{False}$.
\Statex
\Procedure{OnNestedContact}{link, $\mathcal{E}_{contact}$}
    \State $\mathcal{C} \gets \textsc{FilterByTag}(\mathcal{E}_{contact}, \textit{``LeftFingertip''} \lor \textit{``RightFingertip''})$
    \State Update $(c_\ast, f_\ast, a_\ast)$ for the fingertip \textit{link} from $\mathcal{C}$
\EndProcedure
\Statex
\Procedure{Tick}{$\Delta t$}
    \State $F \gets f_R + f_L$;\quad $N \gets c_R + c_L$
    \If{$a_R \neq \varnothing \land a_R = a_L$} \Comment{Both fingertips on same actor}
        \If{$a_R \neq A_{grasp}$}
            \State $A_{grasp} \gets a_R$;\quad \Call{ResetState}{\null}
        \EndIf
        \If{$F > T_{grasp}$}
            \State $\textit{isStable} \gets \text{True}$ \Comment{Stage 1: stable grasp}
            \If{$N < N_{loss}$} \Comment{Stage 2: slip detected}
                \State $\mathcal{I} \gets \mathrm{clip}\!\left(\dfrac{N_{loss} - N}{N_{loss} - N_{min}},\, 0,\, 1\right)$
                 \State \text{TriggerSlipFeedback}($\mathcal{I}$)
            \EndIf
        \Else
            \State \Call{ResetState}{\null} \Comment{Force below grasp threshold}
        \EndIf
    \ElsIf{$A_{grasp} \neq \varnothing$}
        \State $A_{grasp} \gets \varnothing$;\quad \Call{ResetState}{\null} \Comment{Object released}
    \EndIf
    \State Clear per-finger cache $(c_\ast, f_\ast, a_\ast)$
\EndProcedure
\end{algorithmic}
\end{algorithm}

% \begin{itemize}
%     \item We classify essential robot states into two primary categories: \textbf{Confirmation} and \textbf{Exception}.
%     \item \textbf{Confirmations} signal successful completion of sub-tasks, such as "Grasp Stability" or "Complete Insertion".
%     \item \textbf{Exceptions} signal errors or dangers, such as "Linear/Rotational Slip" or "Object Knockover".
%     \item This binary categorization creates a predictable loop of action verification for the operator.
% \end{itemize}

\subsection{Semantic Design}
\label{sec3_semanticdesign}

\begin{figure}
    \centering
    \includegraphics[width=\linewidth]{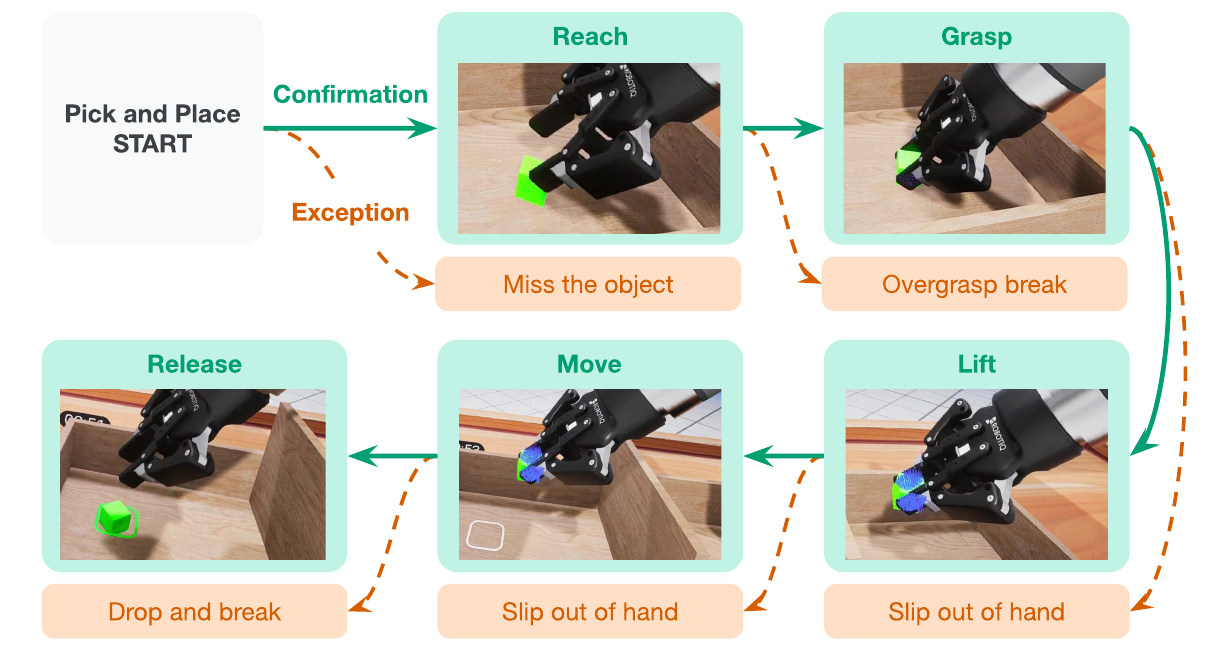}
    \caption{Dexterous manipulation tasks can be partitioned into action phases and subgoals of the overall task.}
    \label{fig:sec3_action_phases}
\end{figure}

Our semantic design draws insights from prior work that investigates how humans utilize tactile signals to perform object manipulation tasks. Johansson and Flanagan~\cite{johansson2009coding} discovered that \textit{"planning and control of manipulation tasks is centred on mechanical events that mark transitions between consecutive action phases and that represent subgoals of the overall task."}. As shown in Figure~\ref{fig:sec3_action_phases}, a simple pick-and-place task can be broken into multiple action phases: reach, load, lift, hold, replace, and unload. The tactile afferents convey important information not only during the action phases, but also to trigger transitions to the next action phase. Therefore, instead of designing tactile feedback that fully replicates the physical sensations of touching objects, our semantic design focuses on \textbf{conveying the completion of action phases (confirmation) and warn the deviation from subgoals (exception)}. For example, when the operator is trying to pick up a fragile object, the semantic design should indicate when the grasp force is enough for pickup and when it is too much that it will break the object.

Because we hypothesize that one of the main advantages of semantic haptic feedback is the generalized one-to-many mapping, we implemented a custom semantic design router in Unreal Engine to enable easy multiplexing between robot state information and haptic feedback design. On one side, it is exposed to the robot state estimation algorithms, allowing them to call functions such as \textit{SendConfirmationMessage()} and \textit{SendWarningMessage()} to convey the essential interaction information to the operator. On the other side, it has control over multiple haptic feedback wearable devices. For each category of message, the task designer can specify the haptic output modality, the stimulation parameters (e.g., amplitude, frequency, duration, etc), and any predefined semantic patterns (e.g., ramp-up, square wave, etc). In this way, state estimation algorithms and haptic pattern design are abstract representations and become independent from each other. Algorithms call functions to send confirmation and exception messages, and the haptic router determines which devices and patterns are used to communicate the messages.

%%% TODO: add a screenshot of the router maybe?

\subsection{Haptic Devices}

% \begin{figure}[htbp]
%     \centering
%     \begin{subfigure}{\columnwidth}
%         \centering
%         \includegraphics[width=\linewidth]{figures/sec3_vibrotactiledesign_v2_20260608.png}
%         \caption{Vibrotactile Feedback Design}
%         \label{fig:sec3_vibrotactiledesign}
%     \end{subfigure}
    
%     \vspace{12pt}
    
%     \begin{subfigure}{\columnwidth}
%         \centering
%         \includegraphics[width=\linewidth]{figures/sec3_pneumaticdesign_v2_20260608.png}
%         \caption{Pneumatic Feedback Design}
%         \label{fig:sec3_pneumaticdesign}
%     \end{subfigure}
    
%     \caption{Multimodal haptic wristbands integrated into the pipeline and their corresponding pattern designs.}
%     \label{fig:sec3_haptic_devices}
% \end{figure}

% \begin{figure*}
%     \centering
%     \includegraphics[width=0.8\linewidth]{figures/sec3_feedbackdesign_v3_20260701.pdf}
%     \caption{Semantic haptic and visual feedback designs used in the evaluation studies for comparison.}
%     \label{fig:sec3_semanticdesign}
% \end{figure*}

Compared to prior haptic teleoperation systems that involve grounded force feedback devices~\cite{chang1999kist, fishel2020tactile} or bulky fingertip tactile devices~\cite{lin2025hcr}, semantic haptic design allows us to build our pipeline upon simple tactile devices such as wristbands. Prior work suggests that the best practice of maximizing the information transmission of haptic devices is to distribute the information across multiple haptic modality and multiple parameters of the same modality~\cite{tan2020methodology}. Therefore, we adopted two tactile modality in our pipeline (Figure \ref{fig:sec3_systempipeline}), a pneumatic wristband adapted from \textit{Bellowband}~\cite{young2019bellowband} and a vibrotactile wristband similar to the one used in ~\cite{xu2024designing}. 

The pneumatic wristband is a soft pneumatic interface composed of six independently controlled thermoplastic polyurethane (TPU) bellows. Unlike bulky grounded systems, this wristband provides a lightweight, low-encumbrance solution for rendering complex haptic cues around the wrist. Each bellow is capable of delivering over 10 N of force and extending beyond 10 mm, allowing the system to provide both high-dimensional local pressure and lower-dimensional cues such as uniform squeezing. With a fast response time under 40 ms and a force bandwidth of 7 Hz, it can deliver both nuanced vibrations and constant force feedback. 
% add the reason why we use 2-4Hz, because Bellowband can only respond up to 7Hz.

The vibrotactile wristband is equipped with four Vybronics VG0840001D linear resonant actuators (LRAs). These compact coin-style motors (8 mm diameter, 1.27 g) feature a fast rise time of 12 ms and a resonance-stabilized vibration force of 1.0 Grms at 170 Hz. Driven by a custom PCB driver, these LRAs provide localized, high-frequency tactile alerts and subtle "tick" sensations that complement the Bellowband's sustained pressure cues.

% TODO: confirm the LRA specs. is it the 0840 LRA from this study? https://docs.google.com/document/d/1ygjxt_A7pQgmFZ7QlCiSp0sVVtHa0AZ-qhiaysNvv4I/edit?tab=t.0
% or the 0825 LRA from this study? https://fb.workplace.com/notes/545651316701697/

\subsection{Haptic Pattern Design}

Following the confirmation-exception task model discussed in Section~\ref{sec3_semanticdesign}, We used wearable devices to convey two types of semantic messages to the operator:
\begin{itemize}
    \item \textbf{Confirmation}: The confirmation message is used to notify the operator when a sub-action is completed, e.g., grasping force is enough to pick up a cube, the USB orientation is aligned with the port; We designed two semantic patterns for this message. The first pattern is a uniform inflation of all the bellows within 0.5 second. This is a bidirectional design, as the bellows can also deflated when the confirmation is no longer valid. The second pattern is a slow ramp-up of vibrations of all actuators within 0.5 second. Vibrations are stopped after the 0.5 period to avoid creating saturated sensations. If the confirmation is no longer valid, a ramp-down vibration will play to notify the user.
    \item \textbf{Exception}: The exception message is used to warn the operator when the current robot movement is deviating away from finishing the sub-action, e.g., grasping force is approaching the threshold of breaking the cube, the USB position is not aligned with the port and about to collide. This message is delivered before the failure event happens, helping the operator actively adjust their input to avoid unwanted event. We designed two semantic patterns for this message. The first pattern is a 5-Hz bellow vibration. The pneumatic controller rapidly opens and closes the valves to achieve this effect. The second pattern is a "Geiger Effect" vibration following the metaphor method~\cite{xu2024designing}. It starts with a slow 2-Hz vibration. As the robot movement is approaching the edge, the vibration frequency gradually increases to 20 Hz, signaling the user of the increased urgency of needing attention.
\end{itemize}

In order to compare the effectiveness of haptic feedback with visual feedback in the evaluation studies, we also implemented a visual feedback design (Figure \ref{fig:sec3_systempipeline}) that added highlight color overlays on top of the gripper. It aims to convey the same messages as semantic haptic feedback but via visual feedback channels.

% \begin{itemize}
%     \item We draw insights from "Tactons" [XXX] —structured tactile messages—to map states to sensations (e.g., a specific pattern for "Create File" vs "Delete File").
%     \item Specific mappings include "Ramp-Up" vibration patterns for grasping confirmation and "Short Pulses" for break warnings.
% \end{itemize}

%% file: docs/4_hapticstudy.tex
\section{Semantic Haptic Comparison Study}
\label{sec4_hapticstudy}

While these haptic designs are backed by common knowledge and prior semantic design methods, it is unknown which combinations of semantic haptic designs will provide the most benefits for operators to perform dexterous manipulation tasks. Thus, we conducted a user study to compare the operator's performance in a pick-and-place task adapted from the classic box and block test~\cite{mathiowetz1985adult}, under various semantic haptic designs. Performance was measured based on quantitative performance metrics and qualitative user feedback.

A total of 12 participants (6 male, 6 female) were recruited for this study. The cohort had a mean age of 33.4 years ($SD = 6.9$). All participants reported regular engagement with haptic devices on a daily or weekly basis. Regarding robotics expertise, eight participants interacted with robots at least monthly, while the remaining four had limited to no prior experience. Each experimental session lasted approximately 60 minutes, and participants were compensated with \$25 USD for their time. This research was approved by the institutional Research Ethics Board.

\subsection{Task Descriptions}

% \begin{figure}
%     \centering
%     \includegraphics[width=\linewidth]{figures/sec4_task_box_and_block_test2.png}
%     \caption{Box and Block test}
%     \label{fig:sec4_task_box_and_block_test}
% \end{figure}

As shown in Figure \ref{fig:sec3_scenes}, Participants are asked to use their right hand to operate a right-hand gripper to perform the adapted box and block test with simulated breakable cubes. The cube is designed to be both slippery and fragile, mimicking the surface property of a piece of soft tofu. If the grip force is too small, the cube will slip out of the gripper; if the grip force is too large, the cube will fracture into pieces. Thus, it requires the participant to utilize the semantic haptic feedback to carefully modulate the grasping force exerted on the cube.

\subsection{Feedback Conditions}

\begin{table}[ht]
\centering
\renewcommand{\arraystretch}{1.3}
\begin{tabular}{l p{3.2cm} p{3.2cm}}
\toprule
\textbf{Condition} & \textbf{Confirmation \newline (Stable Grasp)} & \textbf{Exception \newline (Break Warning)} \\
\midrule
C1 & Vibration & Vibration \\
C2 & Vibration & Air Pressure \\
C3 & Air Pressure & Vibration \\
C4 & Air Pressure & Air Pressure \\
\bottomrule
\end{tabular}
\caption{The four semantic haptic conditions tested in Study One, mapping feedback modalities to confirmation and exception messages.}
\label{tab:haptic_conditions_list}
\end{table}

We conducted a within-subject study, wherein each participant completes four feedback conditions with counterbalanced orders. The four conditions are the 2x2 permutation of semantic messages (grasp confirmation and break warning) and haptic modalities (vibrotactile and air pressure) from Table~\ref{tab:haptic_conditions_list}. Participants wear a VR headset to see the simulation environment and use their hand to control the robot movement. Participants receive haptic feedback when stable grasp is reached or when the grasping force is approaching the cube breaking threshold. 
% without receiving any additional feedback. In analog feedback condition, participants receive pneumatic / vibrotactile feedback on their thumb and index fingertips which are proportionally mapped to the grasping force magnitude computed from the Gripper Force Detector algorithm. In semantic feedback conditions, 

% Our hypotheses are that:
% \begin{enumerate}
%     \item Pneumatic feedback is more suitable for delivering confirmation messages than vibrotactile feedback.
%     \item Vibrotactile feedback is more suitable for delivering exception messages than pneumatic feedback.
%     \item Semantic designs involving multiple haptic modalities are better than those using single modality.
% \end{enumerate}

\subsection{Study Procedure}

At the beginning of the study, the participant reviewed and signed a consent form. The participant was then briefed on the purpose and procedure of the study. And the experimenter helped the participant don all the haptic feedback devices on their right hand and wrist. Following this, the participant experienced the four feedback conditions in a randomized order ($4 \times 4$ Latin square). For each feedback condition, the participant were first asked to go through a practice round by successfully transferring 10 cubes, which often took two to three minutes. Then they were asked to finish the Box and Block test twice, each lasting for one minute. An optional 1-minute break is provided between the two tests. After each condition, participants provided subjective quantitative and qualitative feedback regarding task workload and feedback preference via a questionnaire. After all the conditions were finished, we asked the participant to rank the four conditions and also conducted a short semi-structured interview to collect subjective feedback on the semantic design.
% https://damienmasson.com/tools/latin_square/

\subsection{Performance Metrics}
We evaluate the effectiveness of semantic haptic designs through a combination of teleoperation performance and subjective experience. Primary metrics include the number of cubes transferred, errors (e.g., cubes dropped or broken), average grasp force, and number of grasps per cube transfer. These quantitative measures are complemented by subjective feedback via NASA-TLX questions~\cite{hart1988development}, selected questions from the haptic experience questionnaire~\cite{anwar2023factors}, and teleoperation questions adapted from Khurshid et al.~\cite{khurshid2016effects}:
\begin{itemize}
    \item The feedback was realistic.
    \item The feedback distracted me from the task.
    % \item I like having the feedback as part of the experience.
    \item The feedback reflects varying inputs and events.
    \item How would you describe your awareness of the states of the robot and the object?
    \item How would you describe your level of dexterity while manipulating the object?
\end{itemize}

All data were analyzed using a 2 (Grasp: Vibration vs.\ Air Pressure) $\times$ 2 (Break: Vibration vs.\ Air Pressure) within-subject repeated-measures ANOVA. Normality was assessed via Shapiro--Wilk tests, and sphericity via Mauchly's test; Greenhouse--Geisser corrections were applied when sphericity was violated. For subjective measures where multiple condition cells violated normality, Aligned Rank Transform (ART) ANOVAs \cite{wobbrock2011aligned} were conducted as non-parametric confirmation. Post-hoc pairwise comparisons used Bonferroni correction.

\subsection{Results}

\begin{figure*}[t]
    \centering
    \begin{subfigure}[b]{\textwidth}
        \centering
        \includegraphics[width=\textwidth]{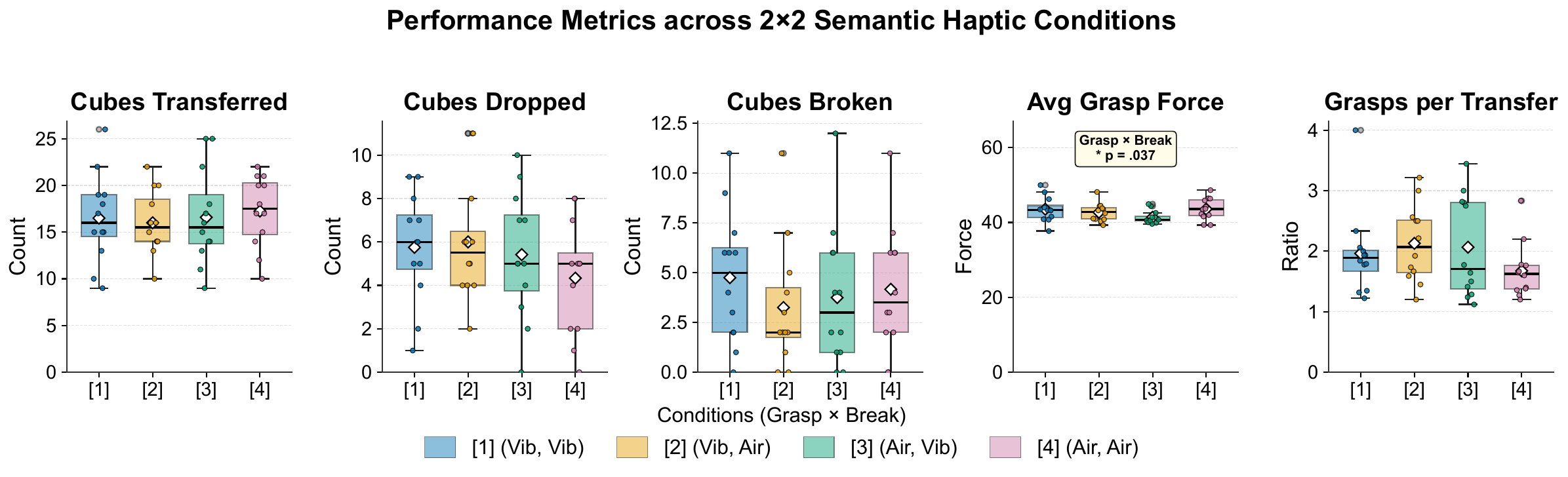}
        % \caption{Study One Performance Metrics.}
        % \label{fig:sec4_studyone_performancemetrics}
    \end{subfigure}

    \vspace{1em}

    \begin{subfigure}[b]{\textwidth}
        \centering
        \includegraphics[width=\textwidth]{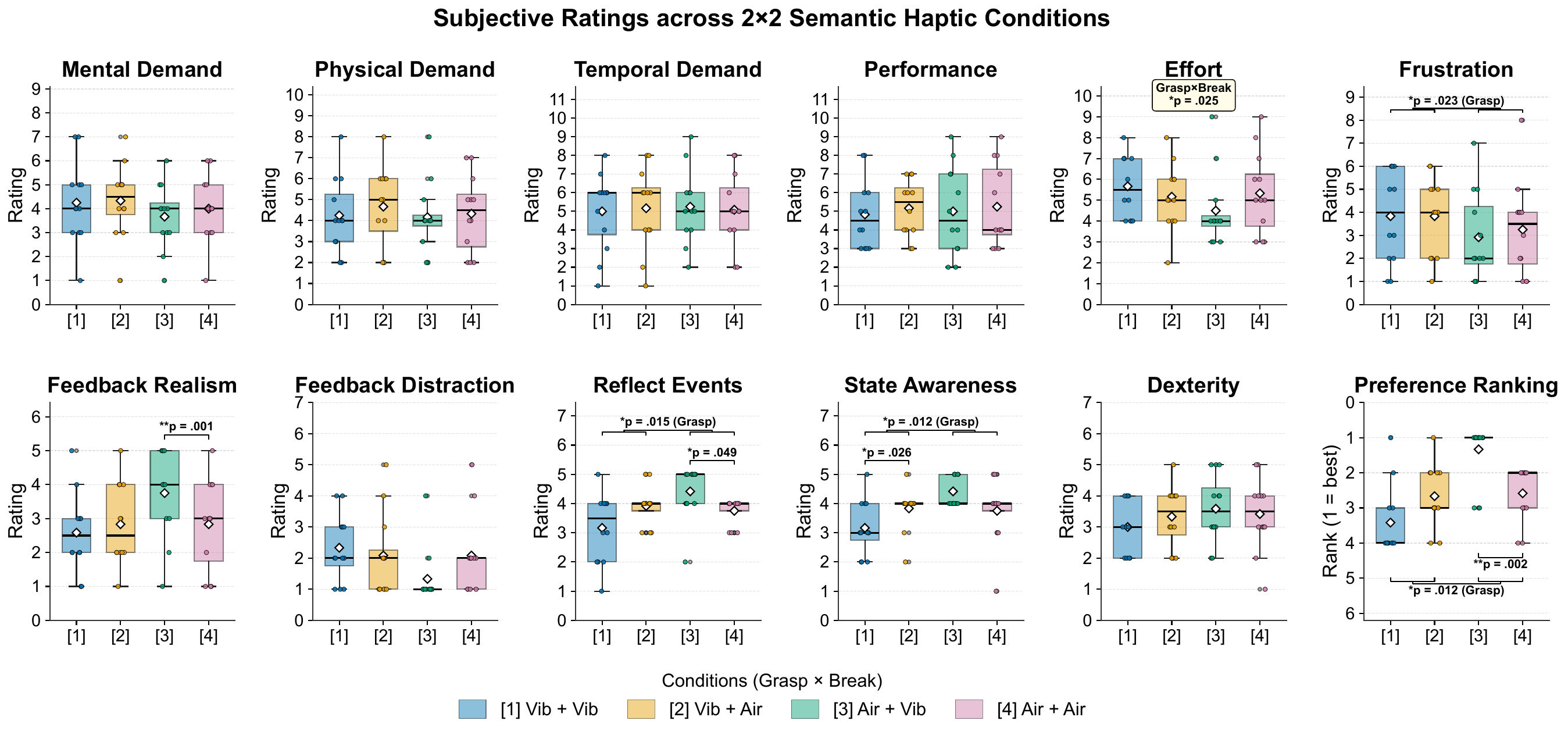}
        % \caption{Study One Subjective Metrics.}
        % \label{fig:sec4_studyone_subjectivemeasures}
    \end{subfigure}
    \caption{Performance metrics and subjective ratings reported in the semantic haptics comparison study. The combination of [Air+Vib] is unanimously preferred by participants, effectively reducing frustration and improving events and state awareness.}
    \label{fig:sec4_studyone_results}
\end{figure*}

\subsubsection{Performance Metrics}

Five performance metrics were evaluated and reported in Fig.~\ref{fig:sec4_studyone_results}.

\textbf{Cubes transferred}: there were no significant differences across feedback conditions (all $p > .50$), indicating that feedback modality did not significantly impact task throughput.

\textbf{Cubes dropped}: no significant effects were found. Although vibrotactile grasp feedback produced numerically more dropped cubes ($M = 5.88$) than pneumatic grasp feedback ($M = 4.88$), this main effect of grasp did not reach significance, $F(1,11) = 3.62, p = .084, \eta^2_p = .247$. Neither the main effect of break feedback, $F(1,11) = 0.33, p = .577$, nor the Grasp $\times$ Break interaction, $F(1,11) = 0.96, p = .348$, was significant.

\textbf{Cubes broken}: no significant effects were found (all $p > .15$), suggesting that haptic feedback modality did not influence the rate of cube breakage.

\textbf{Average grasp force}: a significant Grasp $\times$ Break interaction was observed, $F(1,11) = 5.65, p = .037, \eta^2_p = .339$. However, Bonferroni-corrected post-hoc tests did not reveal significant pairwise differences.

\textbf{Grasps per transfer}: no significant effects were found (all $p > .14$), indicating comparable grasp efficiency across all feedback conditions.

\subsubsection{Subjective Measures}

Subjective responses were collected via a questionnaire comprising six NASA-TLX subscales, three haptic feedback items, two teleoperation items, and a preference ranking (see Fig.~\ref{fig:sec4_studyone_results}).

\textbf{NASA-TLX}: the Raw TLX (RTLX) showed no significant effects (all $p \geq .128$). Individual subscale analyses revealed no significant effects for mental, physical, or temporal demand, or performance (all $p \geq .085$). Regarding \textit{effort}, a significant Grasp $\times$ Break interaction was found, $F(1,11) = 6.77, p = .025, \eta^2_p = .381$; however, no pairwise comparisons survived Bonferroni correction. For \textit{frustration}, a significant main effect of Grasp was observed, $F(1,11) = 6.91, p = .023, \eta^2_p = .386$, with vibration conditions rated higher ($M = 3.83$) than air-pressure conditions ($M = 3.08$).

\textbf{Haptic feedback perception}: for \textit{haptic realism}, a large Grasp $\times$ Break interaction was found, $F(1,11) = 49.00, p < .001, \eta^2_p = .817$. Post-hoc tests revealed that within the air-pressure grasp group, Air+Vib was rated more realistic than Air+Air ($p_{\text{adj}} = .001, d = -0.67$). \textit{Haptic distraction} showed no significant effects (all $p \geq .067$). For \textit{haptic reflects events}, significant effects of Grasp, $F(1,11) = 8.19, p = .015, \eta^2_p = .427$, and interaction, $F(1,11) = 12.67, p = .004, \eta^2_p = .535$, were found, with Air+Vib rated highest.

\textbf{Teleoperation}: \textit{state awareness} showed a significant main effect of Grasp, $F(1,11) = 9.14, p = .012, \eta^2_p = .454$, and interaction, $F(1,11) = 7.65, p = .018, \eta^2_p = .410$. Post-hoc tests indicated higher awareness for Vib+Air than Vib+Vib ($p_{\text{adj}} = .026$). \textit{Dexterity} showed no significant effects (all $p > .120$).

\textbf{Preference ranking}: a significant main effect of Grasp, $F(1,11) = 9.16, p = .012, \eta^2_p = .454$, and a large interaction, $F(1,11) = 22.00, p < .001, \eta^2_p = .667$, were observed. Air+Vib was most preferred ($M = 1.33 \pm 0.78$), significantly outranking Air+Air ($p_{\text{adj}} = .002, d = 1.54$) and Vib+Vib ($p_{\text{adj}} = .021$). Vib+Vib was ranked least preferred ($M = 3.42$).

\subsubsection{Qualitative Feedback}

% Recurring Positives
% Air for stable grasp = realistic, feels like squeezing, continuous reference signal.
% Vibration for break warning = instinctive "alarm" response, quick reaction.
% Distinct modalities for the two events = easier to differentiate without thinking.
% Recurring Negatives
% Issues for using the same modality for both information
%         Vibration for both events (C1) = events blur together.
%         Air break-pulses = sometimes unsettling (P9), too late (P4, P11), can mask grasp pressure (P1).
% Hardware issues: bellow band lag/latency across many sessions; break-warning often arrives after cube already breaks.
% Location of feedback (wrist, not fingers) reduces realism (P7, P9).

To evaluate the user experience and underlying perceptual mechanisms, we conducted a reflexive thematic analysis of the post-experiment interviews.

\textbf{Modality Congruence:} When the haptic modality matched the semantic event, the feedback was intuitive and effective. Steady pneumatic compression for stable grasp felt like an authentic hold; P2 noted that \textit{``with the air when it's compressing, it feels like you're compressing the cube as well. It feels more realistic.''} High-frequency vibration for break warnings elicited an instinctive motor response; P5 reported that \textit{``I view (the vibration) as an alarm unconsciously. When it vibrates, it's more like you're backing up into something and it's telling you that you're gonna hit something, so stop.''}

Conversely, mismatched mappings degraded the experience. Sudden air pulses caused discomfort, with P9 describing that \textit{``the air pulses are really unsettling feeling. It feels like when your heart is beating super fast, but you feel it in your wrist just in a weird way''}. These led users to under-squeeze to avoid the sensation. Vibration for stable grasp created an ongoing buzz that does not link to the grabbing sensation. P2 noted that \textit{``the vibration, your brain doesn't associate that with grabbing something.''}

\textbf{Multimodal Semantic Differentiation:} Separating events into distinct modalities reduced cognitive load. Single-modality designs caused grasp confirmation to be blurred with escalation warnings: P12 noted there is \textit{``very little differentiation between the ramp-up vibration and the high-frequency breaking-point vibration,''} forcing conscious signal evaluation. On contrast, non-overlapping mappings (Air-Stable and Vibration-Break) removed this friction, allowing users to parse success from impending failure without conscious interpretation. P10 praised that \textit{``having two separate haptic sources is more helpful in determining the different events than having it the same type of the same device.''}

\textbf{System Limitations:} Two limitations bounded efficacy. Placing the feedback on the wrist rather than the fingertip could reduce realism but still maintain effectiveness. P7 stated that \textit{``I preferred the haptic sensation to be where I'm actually holding it. But [wrist feedback] definitely helped with the task.''} Latency in the hardware system delayed the warning messages and caused more breaks, as P12 observed that \textit{``I was waiting for the pneumatic haptic to happen when I'm reaching that breaking point. But the breaking object events all happened when I did not receive the pneumatic pulse. There might be a slight delay that could have happened.''}

\subsection{Summary}

The qualitative findings complement the quantitative data and prove \textbf{Air+Vib's dominance} across preference ranking, haptic realism, haptic preference, and ``reflects events''. This is consistent with the modality congruence finding. Steady pneumatic pressure was described as an authentic ``hold'' supplying a continuous reference signal, while high-frequency vibration was characterized as an instinctive ``alarm'' that prompted immediate release. This congruence is further supported by the lower drop rate under air-grasp conditions. The benefit of multimodal semantic differentiation is corroborated by pairwise comparisons on preference ranking, ``reflects events,'' and state awareness, all reflecting the reduced cognitive friction participants described when non-overlapping modalities mapped to events.

Several null effects (e.g., cubes transferred, cubes broken) are likely due to the hardware limitations. Pneumatic latency caused break warnings to arrive slightly after object failure, suggesting that these quantitative gaps could be solidified under improved hardware conditions.

Taken together, the evidence supports adopting Air+Vib (pneumatic grasp confirmation paired with vibrotactile break warning) as the canonical semantic mapping carried forward into the evaluation studies.

%% file: docs/5_unimanualstudy.tex
\section{Unimanual Evaluation With Multimodal Feedback}
\label{sec5_unimanualstudy}

To investigate whether semantic haptic designs have unique strengths over sensory haptic feedback or visual feedback, we conducted a multimodal evaluation study to compare the effects of visual and haptic feedback on teleoperation performance in the same unimanual box-and-block task setting (Figure \ref{fig:sec3_scenes}) used in the comparison study. A total of 12 participants (8 male, 4 female) were recruited for this study. The cohort had a mean age of 34.3 years ($SD = 7.6$). Participants from the previous study were excluded to ensure consistent skill levels across all feedback designs. Regarding prior expertise, eleven participants reported regular (daily or weekly) engagement with haptic devices, while one participant reported rare use. Regarding robotics, eight participants interacted with robots at least monthly, while the remaining four were novices. Each experimental session lasted approximately 60 minutes, and participants were compensated with \$25 USD for their time. This research was approved by the institutional Research Ethics Board.

\subsection{Feedback Conditions}

% \begin{figure}
%     \centering
%     \includegraphics[width=\linewidth]{figures/sec5_studytwo_fourconditions.png}
%     \caption{Four testing conditions used in study two.}
%     \label{fig:sec5_studytwo_fourconditions}
% \end{figure}

\begin{table}[ht]
\centering
\renewcommand{\arraystretch}{1.3}
\setlength{\tabcolsep}{4pt}
\begin{tabular}{@{}l cccc@{}}
\toprule
\textbf{Condition} & \textbf{No Feedback} & \textbf{Visual} & \textbf{Fingertip} & \textbf{Wristband} \\
\midrule
\textbf{Feedback} & \textemdash &
\raisebox{-.5\height}{\includegraphics[width=1.5cm]{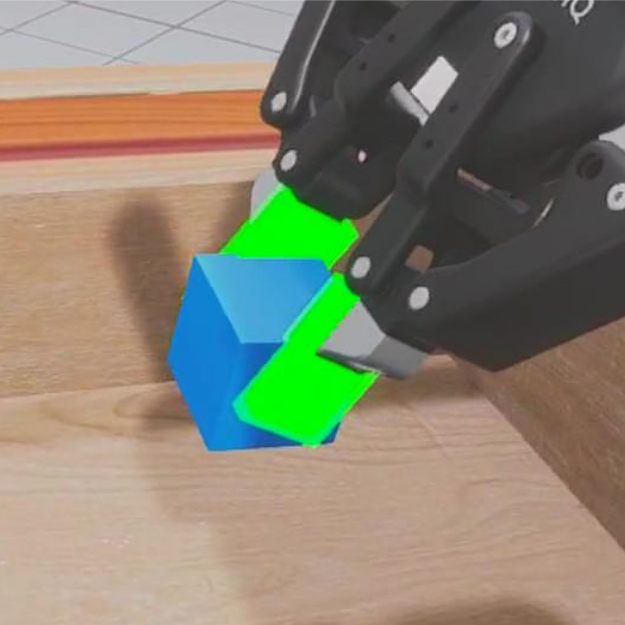}} &
\raisebox{-.5\height}{\includegraphics[width=1.5cm]{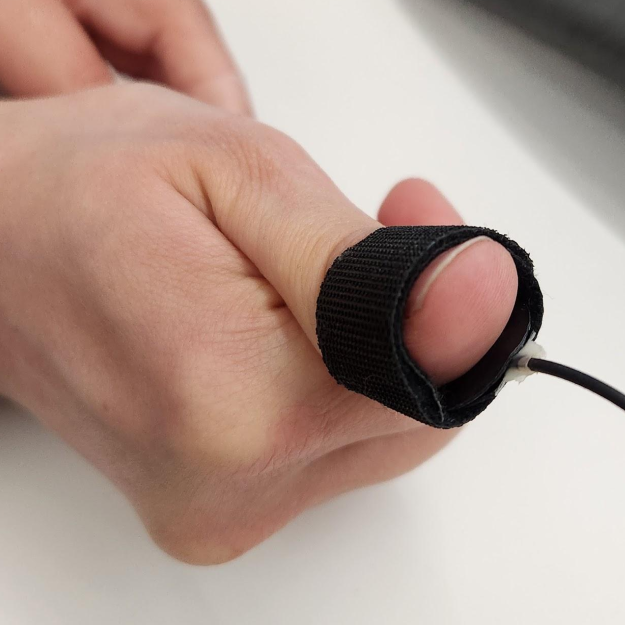}} &
\raisebox{-.5\height}{\includegraphics[width=1.5cm]{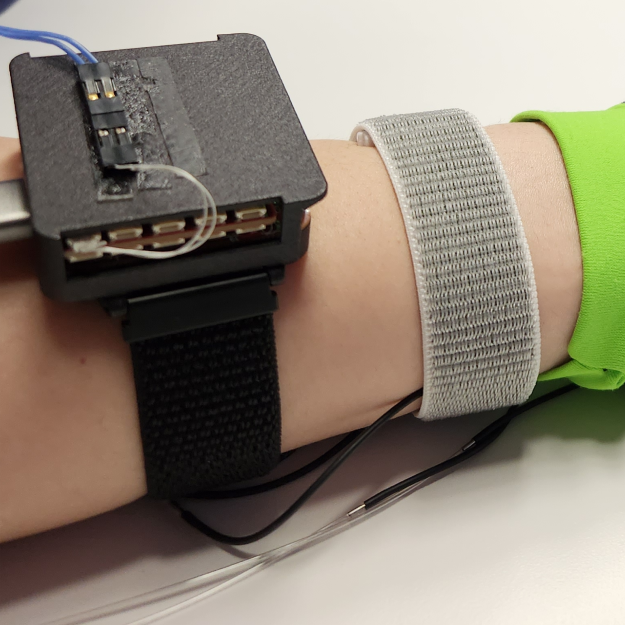}} \\
\bottomrule
\end{tabular}
\caption{Four testing conditions used in study two: no feedback (baseline), semantic visual overlay on the gripper, sensory haptic bellows on the fingertips, and semantic haptic feedback on the wristband.}
\label{tab:sec5_studytwo_fourconditions}
\end{table}

We conducted a within-subject study, wherein each participant completes four feedback conditions with randomly shuffled orders. As shown in Figure \ref{tab:sec5_studytwo_fourconditions}, the four conditions are: no feedback (baseline), semantic visual feedback, sensory haptic feedback and semantic haptic feedback. Semantic visual condition uses the visual overlay design in Figure \ref{fig:sec3_systempipeline}. Sensory haptic condition places bellows on the fingertips of thumb and index fingers, and the air pressure is proportional to the gripper force, aiming to replicate the realistic grasp sensations. Semantic haptic condition utilizes C3 from Table~\ref{tab:haptic_conditions_list} with pneumatic feedback for confirmation and vibrotactile feedback for exception.
% https://ieeexplore.ieee.org/stamp/stamp.jsp?tp=&arnumber=9392357
% ref: Comparison of Three Feedback Modalities for Haptics Sensation in Remote Machine Manipulation

Our hypotheses are that:
\begin{enumerate}
    \item Semantic haptic feedback will improve the number of cubes transferred and reduce errors compared to the other conditions.
    \item Semantic haptic feedback will reduce task workload and improve subjective feedback compared to the other conditions.
    \item Participants will report overall preference of the semantic haptic feedback.
\end{enumerate}

\subsection{Study Procedure}

The procedure for this study followed the same protocol as the previous comparison study, with one key addition to the experimental setup. After the participant donned the wearable devices, including hand tracking gloves, fingertip bellows, and wristbands, these devices remained worn by the participant throughout the entire study to eliminate potential biases induced by varying form factors or wearability.

\subsection{Performance Metrics}

We evaluate the effectiveness of different feedback conditions through the same performance metrics and subjective questions used in the comparison study described in Section~\ref{sec4_hapticstudy}.

All data were analyzed using one-way within-subject repeated-measures ANOVAs across four feedback conditions. Normality was assessed via Shapiro--Wilk tests and sphericity via Mauchly's test. When normality was violated, Friedman tests served as the primary non-parametric analysis. Post-hoc pairwise comparisons used Bonferroni correction (paired $t$-tests for parametric analyses; Wilcoxon signed-rank tests for non-parametric). For multimodal perception items where the No Feedback condition produced floor or ceiling effects, a three-way analysis excluding No Feedback was conducted to assess differences among the active feedback modalities.

\subsection{Results}

\begin{figure*}[t]
    \centering
    \begin{subfigure}[b]{\textwidth}
        \centering
        \includegraphics[width=\textwidth]{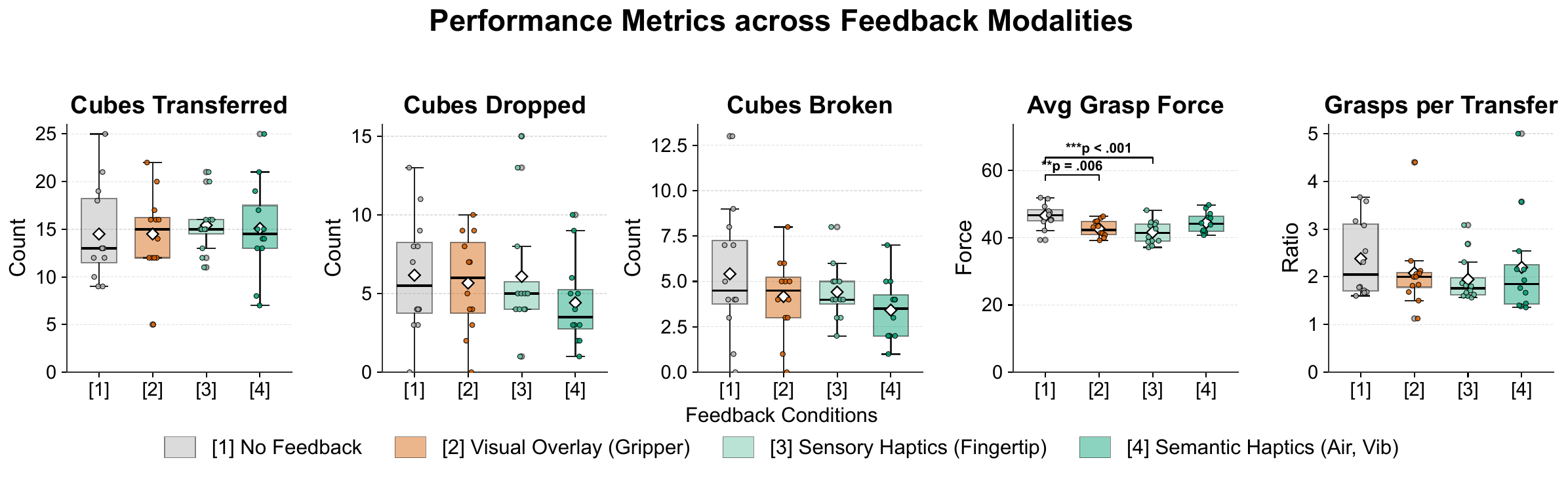}
        % \caption{Study Two Performance Metrics.}
        % \label{fig:sec5_studytwo_performancemetrics}
    \end{subfigure}

    \vspace{1em}

    \begin{subfigure}[b]{\textwidth}
        \centering
        \includegraphics[width=\textwidth]{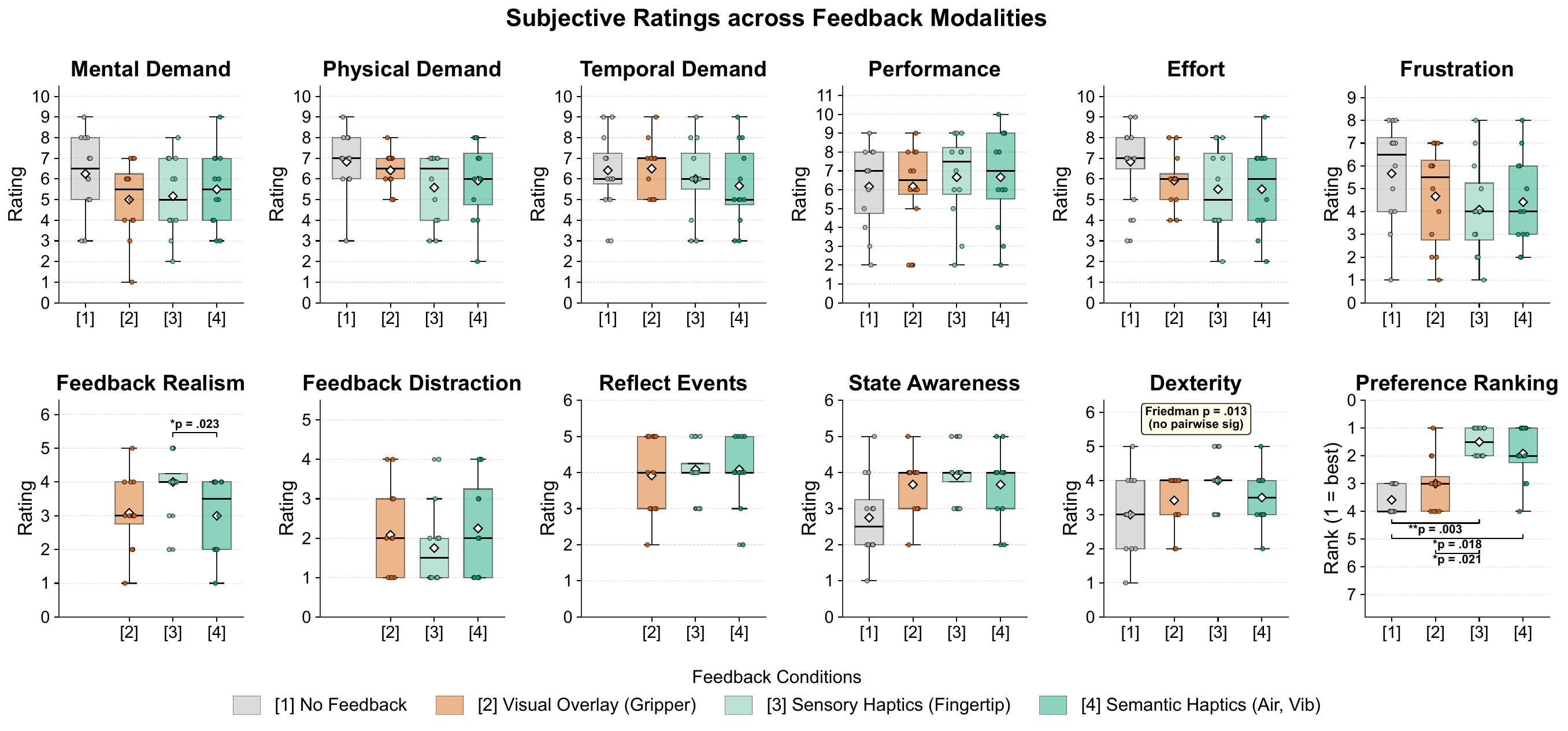}
        % \caption{Study Two Subjective Measures.}
        % \label{fig:sec5_studytwo_subjectivemeasures}
    \end{subfigure}
    \caption{Performance metrics and subjective ratings reported in the unimanual evaluation study. Sensory haptic feedback is mostly preferred by participants due to its high realism and average grasp force reduction.}
    \label{fig:sec5_studytwo_results}
\end{figure*}

\subsubsection{Performance Metrics}

Five performance metrics were evaluated and reported in Fig.~\ref{fig:sec5_studytwo_results}.

\textbf{Cubes transferred}: a one-way RM-ANOVA revealed no significant effect of feedback condition, $F(3, 33) = 0.26, p = .856, \eta^2_p = .023$, indicating that feedback modality did not significantly affect task throughput. % FIX: was p=.805

\textbf{Cubes dropped}: a Friedman test did not reveal a significant effect, $\chi^2(3) = 2.71, p = .439, W = .075$. % FIX: was p=.275

\textbf{Cubes broken}: a one-way RM-ANOVA revealed no significant effect of feedback condition, $F(3, 33) = 1.65, p = .198, \eta^2_p = .130$. % FIX: was p=.334

\textbf{Average grasp force}: a significant main effect of feedback condition was observed, $F(3, 33) = 11.67, p < .001, \eta^2_p = .515$. Bonferroni-corrected post-hoc paired $t$-tests revealed that the No Feedback condition ($M = 46.56 \pm 3.59$) elicited significantly higher grasp forces than the Visual Overlay ($M = 42.71 \pm 2.39, p_{\text{adj}} = .006, g = 1.22$) and Sensory Haptics conditions ($M = 41.62 \pm 3.45, p_{\text{adj}} < .001, g = 1.35$). No other pairwise comparisons reached significance. % FIX: F was 10.57 with df=30; p_adj values and effect sizes updated (effect size is Hedges' g, not Cohen's d)

\textbf{Grasps per transfer}: a Friedman test did not reveal a significant effect, $\chi^2(3) = 3.20, p = .362, W = .089$, indicating comparable grasp efficiency across conditions. % FIX: was p=.277

\subsubsection{Subjective Measures}

Subjective responses were reported in Fig.~\ref{fig:sec5_studytwo_results}.

\textbf{NASA-TLX}: the Raw TLX (RTLX) showed no significant effect of feedback condition, $F(3, 33) = 1.48, p = .237, \eta^2_p = .119$. Individual subscale analyses revealed no significant effects for mental demand ($p = .147$), temporal demand ($p = .378$), performance ($p = .778$), effort ($p = .073$), or frustration ($p = .245$). Physical demand also showed no significant effect via Friedman test, $\chi^2(3) = 6.59, p = .086, W = .183$.

\textbf{Multimodal feedback perception}: analyses excluded the No Feedback baseline due to floor effects (median rating $\approx 1.0$ across the four feedback-specific items). For \textit{feedback realism}, a significant effect was found, $\chi^2(2) = 12.39, p = .002, W = .516$. Bonferroni-corrected Wilcoxon signed-rank post-hoc tests revealed that Sensory Haptics ($M = 4.00$) was rated significantly more realistic than Semantic Haptics ($M = 3.00, p_{\text{adj}} = .023$). No significant effects were observed for \textit{feedback distraction} ($\chi^2(2) = 1.70, p = .428, W = .071$), or \textit{feedback reflects events} ($\chi^2(2) = 0.23, p = .892, W = .010$).

\textbf{Teleoperation}: \textit{state awareness} showed no significant effect, $\chi^2(3) = 6.54, p = .088, W = .182$. For \textit{dexterity}, a significant effect was observed, $\chi^2(3) = 10.77, p = .013, W = .299$; however, no pairwise comparisons survived Bonferroni correction.

\textbf{Preference ranking}: a significant effect was observed, $\chi^2(3) = 19.90, p < .001, W = .553$. Sensory Haptics was ranked most preferred ($M = 1.50 \pm 0.52$), significantly outranking No Feedback ($M = 3.58 \pm 0.51, p_{\text{adj}} = .003$) and Visual Overlay ($M = 3.00 \pm 0.95, p_{\text{adj}} = .021$). Semantic Haptics ($M = 1.92 \pm 1.00$) was also ranked significantly higher than No Feedback ($p_{\text{adj}} = .018$). No significant differences were found between the two haptic conditions or between Semantic Haptics and Visual Overlay. No Feedback was ranked least preferred.

\subsubsection{Qualitative Feedback}

The thematic analysis of post-condition interviews reveals nuanced insights into how participants perceived the type, location, and functional congruence of each feedback condition, and explains the trade-offs underlying the quantitative trends.

\textbf{Co-located Sensory Realism:} The primary advantage of the sensory haptic condition stems from spatial co-location of the feedback at the interaction point. Rendering pneumatic force directly onto the thumb and index fingertips mirrored real-world object interaction and aligned with existing motor mental models, requiring no symbolic translation. As P8 noted, \textit{``Having the feedback on the fingertip is definitely more natural. That's how you expect to have haptic feedback when you pick up stuff.''}

\textbf{Dilemma of Semantic Haptics:} Multimodal feedback on the wrist introduced a clear trade-off. On one hand, rendering the break warning to the wrist resolved a key limitation of sensory haptics, where continuous fingertip pressure alone was difficult to separate a stable grasp from an imminent break event. The vibrotactile channel effectively conveyed proximity to breaking via graded frequency, also outperforming blinking visual warnings; P6 highlighted that \textit{``I really liked when I'm in the red zone the vibration duty cycle would change depending on how close I was to the breaking point, whereas in the earlier study with just the red color, I know I was close but not how close.''} On the other hand, the spatial dislocation pulled focus away from the task space and induced sensory overload, as P10 remarked, \textit{``I was thinking about my wrist more than my fingers when doing it.''}

\textbf{Salient but Distracting Visual Feedback:} The visual overlay provided clear temporal salience and low cognitive load but suffered from spatial distraction. Participants valued its immediate responsiveness, with P7 noting that \textit{``when grasping some objects, the color will immediately turn to green or red.''} However, because the feedback is rendered on the virtual gripper, it forced operators to split visual attention between the target and the robot hand; P8 commented that \textit{``it's a bit distracting because you are supposed to look at the object instead of the robot hand and now I have to pay attention to both.''} This attentional bottleneck caused the feedback to become difficult to use , as P3 stated, \textit{``during the motion I was not able to use the visual cues.''}

% suggest that colocation is a major requirement hmm

\subsection{Summary}

Quantitative and qualitative results converge on a single explanation: sensory haptics' top preference and realism ratings reflect its co-location at the fingertip, while the null effects across performance metrics indicate that any active feedback channel is sufficient for a unimanual box-and-block task whose throughput is bounded by hand motion rather than feedback richness.

Therefore, our hypotheses that semantic haptics would outperform sensory haptics on both performance and preference were not supported in this setting. The wrist-displacement attentional cost offset the semantic clarity gain, leaving semantic haptics statistically comparable to, rather than better than, sensory haptics. This null result motivates the next study: with only one hand and one workspace to attend to, the unique strengths of semantic haptics, such as conveying multiple distinct events without occupying scarce visual or fingertip perception, have little room to manifest. The bimanual task introduced in the following section is designed to expose precisely these attentional and modality-bandwidth constraints.

%% file: docs/6_bimanualstudy.tex
\section{Bimanual Evaluation With Multimodal Feedback}

We extended the study to a \textbf{bimanual pick and place task} to measure the effectiveness of semantic haptic feedback on bimanual teleoperation performance. 

A total of 20 participants (13 male, 7 female) were recruited for this study. The cohort had a mean age of 32.4 years ($SD = 5.1$). 12 of them have attended the previous comparison study. Regarding prior expertise, all participants reported regular engagement with haptic devices on a daily or weekly basis. Regarding robotics, nine participants interacted with robots at least monthly, while the remaining eleven were novices. Each experimental session lasted approximately 70 minutes, and participants were compensated with \$25 USD for their time. This research was approved by the institutional Research Ethics Board.

\subsection{Task Descriptions}

% \begin{figure}
%     \centering
%     \includegraphics[width=\linewidth]{figures/sec6_task_bimanual.png}
%     \caption{Bimanual sorting task}
%     \label{fig:sec6_task_bimanual_sorting}
% \end{figure}

In the virtual environment (Figure \ref{fig:sec3_scenes}), participants are asked to use both of their hands to control grippers to finish a bimanual sorting task. More specifically, the left gripper is in charge of picking and holding a square container, with an opening on the top for cube insertion. The right gripper is in charge of picking up cubes and dropping them into the container. The cube is the same as the one used in the box and block test, with preset grasp and break threshold. The container is designed with low friction coulomb coefficients so that even though participants grasp it with sufficient force, it will slip out of the gripper every 10-15 seconds. When the slip is happening, the operator is instructed to put it back to the base and adjust the gripper pose. If the container drops to the table, it will respawn after a two-second penalty interval. Thus, if no additional feedback is provided during the task, the operator needs to regularly switch their attention back and forth between the two objects to make sure no slippage or breaking is about to happen. The left gripper is equipped with the Gripper Slip Detector script to assist with slip detection.

\subsection{Feedback Conditions}

We conducted a within-subject study, wherein each participant completes four feedback conditions with randomly shuffled orders. The four conditions are: no feedback (baseline), semantic visual feedback, sensory haptic feedback and semantic haptic feedback. For all four conditions, the feedback provided to the right hand is the same as the unimanual study, while the feedback provided to the left hand shares similar designs but focuses on the slippage warning. For example, in visual feedback condition, participants would see visual overlay of blinking red light for slippage warning. In the sensory haptic condition, air pressure is linearly mapped to the grasp force on the container. In semantic haptic condition, pneumatic wristband is worn on the left arm to provide grasp confirmation and vibrotactile wristband provides slippage warning.

Our hypotheses are the same ones used in the unimanual study in Section~\ref{sec5_unimanualstudy}, hypothesizing that semantic haptic feedback will improve teleoperation performance, reduce workload and receive overall preference.
% \begin{enumerate}
%     \item Semantic haptic feedback will improve the number of cube transferred and reduce errors compared to the other conditions.
%     \item Semantic haptic feedback will reduce task workload and improve subjective feedback compared to the other conditions.
%     \item Participants will report overall preference of the semantic haptic feedback.
% \end{enumerate}

\subsection{Study Procedure}

The experimental procedure followed the same workflow as the previous study, adapted for a bimanual setup. Participants donned the tracking gloves, fingertip bellows, and wristbands on \textbf{both hands} and completed a \textbf{bimanual pick-and-place test}. Each of the two tests per condition lasted for \textbf{two minutes} (up from one minute). The remaining procedural elements—including the consent process, the 10-cube practice round, the optional 1-minute break, and the sequence of questionnaires, ranking, and interviews—remained identical to the previous protocol.

\subsection{Performance Metrics}

We evaluate the effectiveness of different feedback conditions through similar performance metrics and subjective questions as the ones used in the unimanual study described in Section~\ref{sec5_unimanualstudy}. A few performance metrics are added to evaluate bimanual interactions, such as the number of containers dropped onto the table, average peak grasping force, and average number of cube transfers per container. Subjective questions remain the same. All data were analyzed using the same analyses as the unimanual study.

\subsection{Results}

\begin{figure*}[t]
    \centering
    \begin{subfigure}[b]{\textwidth}
        \centering
        \includegraphics[width=\textwidth]{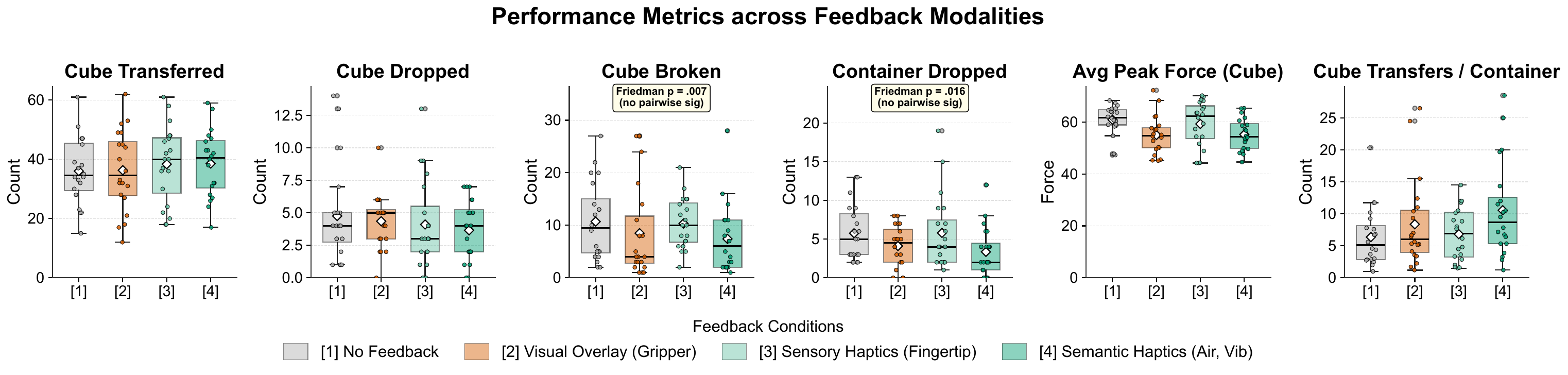}
        % \caption{Study Three Performance Metrics (Bimanual Task).}
        % \label{fig:sec6_studythree_performancemetrics}
    \end{subfigure}

    \vspace{1em}

    \begin{subfigure}[b]{\textwidth}
        \centering
        \includegraphics[width=\textwidth]{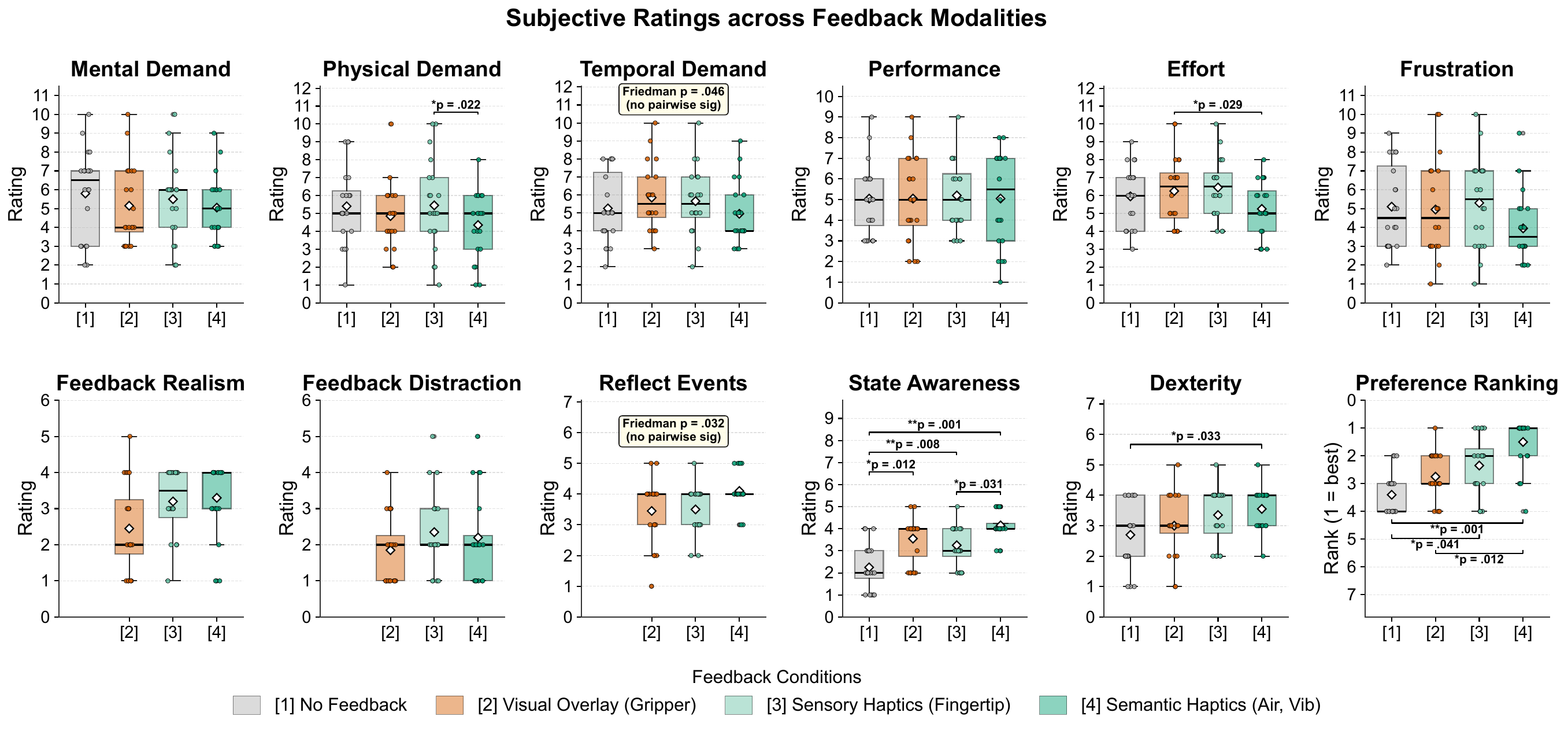}
        % \caption{Study Three Subjective Measures.}
        % \label{fig:sec6_studythree_subjectivemeasures}
    \end{subfigure}
    \caption{Performance metrics and subjective ratings reported in the bimanual evaluation study. Semantic haptic feedback is found to be most effective by reducing physical demand and effort while improving state awareness and dexterity.}
    \label{fig:sec6_studythree_results}
\end{figure*}

\subsubsection{Performance Metrics}

Six performance metrics were evaluated for the bimanual task and are reported in Fig.~\ref{fig:sec6_studythree_results}.

\textbf{Cubes transferred}: a one-way RM-ANOVA revealed no significant effect of feedback condition, $F(3, 57) = 0.82, p = .489, \eta^2_p = .043$, indicating that feedback modality did not affect bimanual task throughput.

\textbf{Cubes dropped}: a Friedman test did not reveal a significant effect, $\chi^2(3) = 3.03, p = .387, W = .051$.

\textbf{Cubes broken}: a Friedman test revealed a significant effect of feedback condition, $\chi^2(3) = 12.17, p = .007, W = .203$, with No Feedback ($M = 10.70 \pm 7.49$) and Sensory Haptics ($M = 10.35 \pm 4.84$) showing the highest counts and Semantic Haptics the lowest ($M = 7.40 \pm 6.60$). However, no pairwise comparisons survived Bonferroni correction (all $p_{\text{adj}} \ge .18$). % FIX: borderline ".18" -> ">= .18" (actual min .182)

\textbf{Containers dropped}: a Friedman test revealed a significant effect of feedback condition, $\chi^2(3) = 10.27, p = .016, W = .171$, with Semantic Haptics yielding the fewest drops ($M = 3.35 \pm 3.17$) and Sensory Haptics the most ($M = 5.80 \pm 4.80$). No pairwise comparisons survived Bonferroni correction (all $p_{\text{adj}} \ge .08$). % FIX: was "> .07" (actual min .078)

\textbf{Average peak grasp force (right hand, cube)}: a Friedman test did not reveal a significant effect, $\chi^2(3) = 6.84, p = .077, W = .114$.

\textbf{Average cube transfers per container cycle}: a Friedman test did not reveal a significant effect, $\chi^2(3) = 6.18, p = .103, W = .103$.

\subsubsection{Subjective Measures}

Subjective responses are reported in Fig.~\ref{fig:sec6_studythree_results}. % FIX: removed stray ")"

\textbf{NASA-TLX}: the Raw TLX (RTLX) showed no significant effect of feedback condition, $F(3, 57) = 2.11, p = .109, \eta^2_p = .100$. Individual subscale analyses revealed no significant effects for mental demand ($p = .131$), performance ($p = .773$), or frustration ($p = .112$). % FIX: replaced "all p > .11" with explicit values
For physical demand, a significant main effect was observed, $F(3, 57) = 3.37, p = .025, \eta^2_p = .150$; Bonferroni-corrected paired $t$-tests revealed that Sensory Haptics ($M = 5.45 \pm 2.54$) was rated significantly more physically demanding than Semantic Haptics ($M = 4.35 \pm 1.87, p_{\text{adj}} = .022, g = -0.48$). For temporal demand, a significant omnibus effect was found, $\chi^2(3) = 7.99, p = .046, W = .133$, but no pairwise comparisons survived Bonferroni correction. For effort, a significant effect was observed, $\chi^2(3) = 10.52, p = .015, W = .175$; Bonferroni-corrected Wilcoxon signed-rank post-hoc tests showed that Semantic Haptics ($M = 5.25 \pm 1.55$) required significantly less effort than Visual Overlay ($M = 6.25 \pm 1.94, p_{\text{adj}} = .029$).

\textbf{Multimodal feedback perception}: analyses excluded the No Feedback baseline due to floor effects (median rating $\approx 1.0$ across the four feedback-specific items). No significant effects were found for feedback realism ($\chi^2(2) = 5.30, p = .071, W = .133$) or feedback distraction ($\chi^2(2) = 1.03, p = .597, W = .026$). For feedback reflects events, a significant omnibus effect was found, $\chi^2(2) = 6.86, p = .032, W = .172$, but no pairwise comparisons survived Bonferroni correction (all $p_{\text{adj}} \ge .05$). % FIX: was "> .05" (actual min .051)

\textbf{Teleoperation}: state awareness showed a significant effect, $\chi^2(3) = 30.43, p < .001, W = .507$. All three active modalities were rated significantly higher than No Feedback ($M = 2.25 \pm 1.02$, all $p_{\text{adj}} \le .012$), and Semantic Haptics ($M = 4.15 \pm 0.59$) was rated significantly higher than Sensory Haptics ($M = 3.25 \pm 0.91, p_{\text{adj}} = .031$). For dexterity, a significant effect was found, $\chi^2(3) = 14.01, p = .003, W = .233$, with Semantic Haptics ($M = 3.55 \pm 0.69$) rated significantly higher than No Feedback ($M = 2.70 \pm 1.08, p_{\text{adj}} = .033$); no other pairwise comparisons survived correction.

\textbf{Preference ranking}: a significant effect was observed, $\chi^2(3) = 22.74, p < .001, W = .379$. Semantic Haptics was ranked most preferred ($M = 1.50 \pm 0.89$), significantly outranking both No Feedback ($M = 3.40 \pm 0.75, p_{\text{adj}} = .001$) and Visual Overlay ($M = 2.75 \pm 0.85, p_{\text{adj}} = .012$). Sensory Haptics ($M = 2.35 \pm 1.09$) was also ranked significantly higher than No Feedback ($p_{\text{adj}} = .041$). No significant differences were found between the two haptic conditions or between Sensory Haptics and Visual Overlay. No Feedback was ranked least preferred.

\subsubsection{Qualitative Feedback}

While co-located sensory haptics were preferred for unimanual operations, adding a second active hand shifted preference toward wrist-based semantic haptic feedback. Thematic analysis indicates this reversal is driven by the modality limits and attentional demands.

\textbf{Modality Limits of Sensory Haptics:} While physically realistic, continuous fingertip feedback proved less effective for dual-task exception handling. Because this modality strictly renders realistic sensations, the continuous pneumatic pressure cannot convey the shear force induced by slip; P8 explained that \textit{``you don't really notice it's slipping until it's almost too late to correct it.''} Furthermore, under high cognitive load, the constant signal is prone to sensory adaptation, as P14 noted \textit{``initially I feel the air bubble inflates, but over time it doesn't call attention to me at all...So I have to focus on it to understand there is tactile feedback.''}

\textbf{Attentional Collapse of Visual Feedback:} Under bimanual load, visual feedback was limited by the operator's field of view: with visual attention typically dedicated to fine manipulation on one hand, the stabilizing hand was often left unmonitored. P3 described this bottleneck, noting that \textit{``because when I'm doing this manipulation I keep my eyes to the right hand...I don't notice the color of the back hand.''} Furthermore, when exceptions occurred on both grippers simultaneously, the concurrent visual alerts increased cognitive stress; P15 described that they were \textit{``most stressed when both are blinking red and it just feels so unnecessarily urgent...this isn't that serious of a task but it would like stress me out.''}

\textbf{Eyes-Free Distinct Messages via Semantic Haptics:} Semantic wrist feedback addressed these limitations by providing discrete guidance that does not compete for visual gaze, allowing operators to monitor the secondary hand without looking away from the primary task. P10 praised this capability, noting that \textit{``I could even do that without looking. So it was very useful because it solved the vision problem.''}. Spatially separating the feedback across the two wrists also helped operators manage concurrent tasks; P4 described \textit{``if I was picking up the right block and I got the left block signal, I think it was clear to me that I needed to put it down.''} A minor limitation involved cross-hand interference, where an abrupt alert could disrupt the opposing task, as P13 observed that \textit{``sometimes I break the cube because the cup starts vibrating.''}

\subsection{Summary}

Quantitative and qualitative results converge to support our hypotheses in the bimanual setting. Semantic haptics was top-ranked in preference, rated highest on state awareness, and required less physical demand than sensory haptics and less effort than visual feedback. These advantages align directly with the eyes-free, spatially distinct nature of wrist feedback, which freed visual attention for the primary hand while keeping the stabilizing hand monitored through an independent sensory system. The reduction in containers dropped and cubes broken is consistent with participants reporting that the vibrotactile slip warning cut through the adaptation and attentional bottlenecks that limited the other two feedback conditions.

The performance hypothesis was only partially supported: although the trends favored semantic haptics on error metrics, no pairwise comparisons survived Bonferroni correction. We attribute this to the within-subject variability of bimanual coordination strategies, which can mask modality-level effects on count-based outcomes. Overall, the bimanual results validate the intended strengths of semantic haptics: discrete, modality-distinct messages on spatially separated body sites scale more gracefully than co-located sensory haptics once attentional and hardware bandwidth become the binding constraints.

%% file: docs/7_discussion.tex
\section{Discussion}

Based on the findings from evaluation studies, we summarize the strengths of semantic haptic feedback, extract design guidelines, and discuss limitations of the current implementation and potential future work.

\subsection{Strengths of Semantic Haptic Feedback}

Across our three evaluation studies, we investigated how semantic haptic feedback impacts teleoperation performance under various design choices and task configurations. The initial comparison study identified the optimal semantic design among four factorial configurations, while the subsequent unimanual and bimanual studies highlighted when and why semantic feedback outperforms state-of-the-art alternatives and remains preferred by users.

\textbf{Haptic modality congruence} is fundamental to effective semantic design. Our comparison study demonstrated that pneumatic feedback is more suited for grasp confirmation due to its steady pressure state, whereas vibrotactile feedback is better optimized for warnings by mimicking alarm sensations. This aligns with our design rationale of leveraging real-world metaphors and affective touch; high-frequency vibrations evoke intuitive warning responses (similar to automotive proximity beeps~\cite{van2004vibrotactile} or smartwatch alerts~\cite{brown2006feel}), prompting participants to loosen their grip to prevent cube breakage. Even though pneumatic feedback is applied to the wrist rather than the fingertip interaction point, its steady pressure provides a reassuring sense of safety, which is consistent with findings in affective touch research~\cite{yohanan2012role}. Aligning haptic modalities with a user's internal mental model ultimately reduces frustration while enhancing both realism and state awareness.

\textbf{One-to-many mapping} allows a unified semantic haptic framework to generalize across diverse task requirements. In our bimanual study, pneumatic feedback consistently signaled grasp confirmation for both cubes and containers across both wrists. Meanwhile, vibrotactile feedback successfully conveyed two distinct warning messages that demanded opposite user responses: a cube-breaking warning (requiring loosening the grip) and a container-slipping warning (requiring tightening or adjusting the grip). Despite this duality, participants reacted quickly and without confusion. In contrast, conventional sensory feedback utilized identical fingertip pressure for both states and failed to convey incipient slip altogether. Articulating slip realistically through sensory feedback would require complex, cost-prohibitive shear-force hardware. This suggests that semantic haptic feedback can elegantly scale a single hardware footprint to multiple tasks through one-to-many mappings.

\textbf{Simplified hardware} decouples haptic rendering from the exact physical interaction point, significantly expanding the design space for wearable form factors. Traditional sensory haptics demand colocated feedback (e.g., fingertip actuators for a pick-and-place task). However, scaling a robotic system from simple grippers to fully dexterous hands introduces severe hardware constraints, as actuators must then be distributed across numerous contact sites like fingers, joints, and palms. In contrast, semantic haptics relaxes this positional constraint. For instance, a single pneumatic wristband can communicate grasp confirmation for arbitrary end-effectors and object geometries, greatly reducing hardware complexity and enhancing system reusability.

\textbf{Reduced workload and improved awareness} lead to overall preference. Semantic haptic feedback provides clearly distinguishable cues to the participant so they can form a clear mental model of the current states between robots and objects. As comparison, sensory haptic feedback requires participants to mentally estimate whether they have exert enough force to pick up the cube or too much force to break it, which causes significantly higher physical demand. In bimanual settings, the advantage is even more apparent, Compared to semantic visual feedback, semantic haptic feedback enables participants to simultaneously track the grasp states of both grippers without needing to frequently turning their head, thus reducing efforts and being more preferred. 

\textbf{Dimension reduction}. Fundamentally, what semantic haptic has demonstrated in this work, is the ability to convert high-dimensional contact-rich information in dexterous manipulation tasks to low-dimensional haptic semantic messages conveyed via wearable devices. It suggests a new path towards haptic assisted teleoperation design: instead of trying to replicate the realistic sensations of multimodal touch interaction, use touch as an intuitive communication channel to convey the most essential robot state to the operator so they can perform dexterous tasks more confidently.

\subsection{Design Guidelines}

Our semantic haptic pipeline and evaluation studies served as an initial exploration of the enormous design space of semantic haptic feedback for robot teleoperation. There are many other haptic rendering techniques and dexterous manipulation tasks for other researchers to explore. To help accelerate their exploration process, we summarized several design guidelines:

\begin{itemize}
    \item \textbf{Identify the most important object state information} for a certain task. This is crucial because we can't use haptic feedback to tell the operator everything about the interaction. Thus, we need to carefully choose the ones that are important for each task. Regarding the two tasks chosen in this work, we identify the most important state information as grasp confirmation, break warning, and slip exception. But for other tasks, this could be entirely different. Researchers are encouraged to breakdown tasks using the multi-stage theory and identify the most essential info for each stage.
    \item \textbf{Leverage generalized semantic mapping} to simplify hardware design. If the semantic meaning is similar across multiple stages of one task or multiple tasks, the same haptic design can be reused without worrying about causing confusion. For example, both grasp confirmation in pick-and-place tasks and orientation alignment in insertion tasks can be treated as "positive confirmation" messages and communicated with a firm squeeze on the wrist using pneumatic wristbands.
    \item \textbf{Hands-on experience} accelerates the design process. In our design iterations, we have collected many semantic patterns from literature that claim to work very well, but in reality when they were tested in teleoperation, they were either too complex and difficult to interpret during rapid interaction, or too naive and cannot communicate the necessary info regarding the robot state. It takes a lot of trial-and-error to find the right designs that balance in both axes. This also motivates us to build the haptic router module so we can easily test different haptic patterns without heavily modifying the codes. Thus, we suggest that once researchers have a rough idea of what haptic hardware and patterns can be effective for their task, it is beneficial to test them directly in the task rather than keep polishing the design as a sole haptic effect.
\end{itemize}

\subsection{Limitations and Future Work}

Although we demonstrate the first semantic haptic teleoperation pipeline, several limitations remain.

First, integrating semantic haptics requires a priori knowledge of object properties (e.g., breaking force) and precise state estimation (e.g., slip detection); we simplify this with a simulation where properties are known and contact force is derived from real-time physics. Extending to physical robots could use vision-language models to recognize objects and estimate their properties and tactile sensors on end-effectors to estimate grasp or slip states.

Second, our design uses a single pattern per modality (constant pressure, high-frequency vibration), which suffices for pick-and-place but not for tasks needing complex transitions or continuous tracking. Scaling will require multiple patterns within a modality while maintaining perceptual distinctiveness, which future work could address via Multidimensional Scaling to design maximally differentiable haptic icons~\cite{maclean2003perceptual}.

Third, some participants noted that the spatial displacement between the interaction site (fingertip) and feedback site (wrist) increased cognitive load, so evaluating fingertip-rendered semantic feedback is a valuable direction.

Finally, our evaluation only used parallel-jaw grippers. Extending to multi-fingered dexterous hands should be straightforward on the hardware side, with the main challenge being robust state estimation for the complex hand-object interactions of higher-DoF end-effectors.

\subsection{Demonstration of Use Cases}

Although this work evaluates semantic haptic feedback only on pick-and-place teleoperation, the approach generalizes to other dexterous manipulation tasks. We illustrate two additional use cases we explored with the same pipeline.

\textbf{USB insertion} assesses unilateral position and orientation alignment. The operator picks up a USB-A connector and inserts it into a port placed at a randomized pose. A state estimator continuously compares the connector and port transforms: a ramp-up pneumatic cue confirms rotational alignment, while vibrotactile feedback warns of imminent positional collisions, guiding the operator toward a smooth insertion.

\textbf{Dish wiping} assesses bimanual force modulation on fragile objects. The operator wipes stains from a ceramic plate using a sponge held by an end effector with variable impedance, of which stiffness changes are not visible to the operator. Pneumatic pressure ramps up upon contact to signal engagement, and continuous vibration renders the wiping force measured by an F/T sensor, warning the operator when the applied force approaches unsafe levels.

\subsection{Conclusion}

We introduced semantic haptic feedback for dexterous teleoperation, conveying abstract robot states as confirmations and exceptions through a reconfigurable pipeline with pneumatic and vibrotactile wristbands. Through a series of user studies, we identified the best semantic haptic design and demonstrated that in a bimanual pick-and-place study, semantic haptics reduced workload, improved situational awareness, and was the most preferred condition. These results suggest that abstracting high-dimensional contact information into low-dimensional, modality-congruent messages is a scalable direction for haptic-assisted teleoperation.